\documentclass[usenames,dvipsnames]{article}

\usepackage[utf8]{inputenc}
\usepackage{glossaries} 
\usepackage[colorlinks=true,allcolors=blue]{hyperref}
\makeglossaries
\usepackage[french, english]{babel} 

\usepackage{xspace}

\usepackage{natbib} 
\usepackage{graphicx}
\usepackage[usenames,dvipsnames]{xcolor} 
\graphicspath{{./}}

\usepackage{algorithm}
\usepackage{algorithmic}

\usepackage{amsmath} 
\usepackage{mathrsfs} 
\usepackage{amsthm} 
\usepackage{amsfonts} 
\usepackage{amssymb} 
\usepackage{dsfont} 
\usepackage{stmaryrd} 

\usepackage[colorinlistoftodos,textwidth=2.3cm]{todonotes}

\newcommand \ie {i.e.\xspace}

\newcommand \vs {{vs}\xspace}

\newtheorem{theorem}{Theorem}
\newtheorem{lemma}[theorem]{Lemma}

\newtheorem{property}[theorem]{Property}

\newtheorem{remark}{Remark}[section]

\newcommand \R {\mathbb{R}\xspace}
\newcommand \N {\mathbb{N}\xspace}
\newcommand \set[1] {\mathcal{#1}}
\newcommand \X {\set{X}\xspace}

\newcommand \F {\set{F}\xspace}

\newcommand \Y {\set{Y}\xspace}

\newcommand \lin {\operatorname{span}}
\newcommand \prox {\operatorname{prox}}
\newcommand \sign {\operatorname{sign}}

\newcommand{\iiprod}[3]{\left\langle #2, #3 \right\rangle _ {#1}}
\newcommand{\iprod}[2]{\iiprod{}{#1}{#2}}
\newcommand{\nnorm}[2]{\left\| #2 \right\| _ {#1}}
\newcommand{\norm}[1]{\nnorm{2}{#1}}

\newcommand \e {\operatorname{e}}

\newcommand{\prob}{\mathbb{P}}

\newcommand{\expe}{\operatorname{\mathbb{E}}}

\usepackage{amsmath}

\newcommand{\argmin}{\operatorname{arg \, min}}

\newcommand \minimize[1] {\underset{#1}{\operatorname{minimize}}~}

\newenvironment{disarray}
 {\everymath{\displaystyle\everymath{}}\array}
 {\endarray}
\newenvironment{opb*} 
{
	\[
		\begin{disarray}{c@{\hspace*{0.05cm}}l}
}
{
		\end{disarray}
	\]\hspace*{-1.1ex}
}
\newcounter{opb}
\renewcommand*{\theopb}{P\arabic{opb}}

\makeatletter
\renewcommand*{\p@opb}[1]{
  (#1)
}
\makeatother

\newenvironment{opb}[1]
{
	\refstepcounter{opb}
	\label{#1}
	\equation
		\begin{disarray}{c@{\hspace*{0.05cm}}l}
}
{
		\end{disarray}
		\tag{\theopb}
	\endequation
}

\newcommand \fgrad {\nabla_n}
\newcommand \fsub {\widetilde \nabla_n}
\newcommand \fset{\mathscr F_{\X}}
\newcommand \proxdir[1]{\operatorname{Prox}_n^{#1}}

\usepackage{ulem}
\newcommand{\replace}[2]{#2}

\newcommand{\new}[1]{\textbf{{\color{Green} #1}}}
\renewcommand{\new}[1]{#1}
\newcommand{\neurocomp}[1]{\textbf{{\color{Green} #1}}}
\renewcommand{\neurocomp}[1]{#1}

\usepackage{enumitem}
\newenvironment{assumptions}
 {\enumerate[label=\textbf{(A\arabic*)}, ref=(A\arabic*), align=left, leftmargin=20pt]}
 {\endenumerate}
\makeatletter
\newcommand\hyp[1]{\item[\textbf{(#1)}]
  \edef\@currentlabel{(#1)}}
\makeatother
\makeatletter
\newcommand\varitem[1]{\item[\textbf{(A\arabic{enumi}{$#1$})}]
 \edef\@currentlabel{(A\arabic{enumi}{$#1$})}}
\makeatother

\newcommand{\numberthis}{\addtocounter{equation}{1}\tag{\theequation}}

\newcommand{\myacronym}[4] 
{
	\newglossaryentry{#1}
	{
		type=\acronymtype,
		name={#2},
		description={#3},
		text={#2},
		first={#3 (#2)},
		firstplural={#4 (#2s)},
		short={#2}
	}
	\expandafter\newcommand\csname #1\endcsname{\gls{#1}\xspace} 
	\expandafter\newcommand\csname #1s\endcsname{\glspl{#1}\xspace} 
}

\myacronym{rkhs}{RKHS}{reproducing kernel Hilbert space}{reproducing kernel Hilbert spaces}
\myacronym{ovk}{OVK}{operator-valued kernel}{operator-valued kernels}
\myacronym{svm}{SVM}{support vector machine}{support vector machines}
\myacronym{svc}{SVC}{support vector classification}{-}
\myacronym{svr}{SVR}{support vector regression}{-}
\myacronym{krr}{KRR}{kernel ridge regression}{-}
\myacronym{mmr}{MMR}{maximum margin robot}{maximum margin robots}

\myacronym{lda}{LDA}{linear discriminant analysis}{Linear discriminant analysis}
\myacronym{qda}{QDA}{quadratic discriminant analysis}{Quadratic discriminant analysis}
\myacronym{samme}{SAMME}{Stagewise Additive Modeling using a Multiclass Exponential loss}{-}
\myacronym{dbscan}{DBSCAN}{density-based spatial clustering of applications with noise}{-}
\myacronym{bic}{BIC}{Bayesian information criterion}{Bayesian information criteria}
\myacronym{aic}{AIC}{Akaike information criterion}{Akaike information criteria}
\myacronym{pca}{PCA}{principal component analysis}{-}
\myacronym{svd}{SVD}{singular value decomposition}{-}
\myacronym{rip}{RIP}{restricted isometry property}{-}
\myacronym{mds}{MDS}{multidimensional scaling}{-}
\myacronym{smacof}{SMACOF}{scaling by majorizing a complicated function}{-}

\myacronym{mst}{MST}{minimum spanning tree}{minimum spanning trees}

\myacronym{admm}{ADMM}{alternating direction method of multipliers}{-}
\myacronym{smo}{SMO}{sequential minimal optimization}{-}
\myacronym{kkt}{KKT}{Karush-Kuhn-Tucker}{-}
\myacronym{pdcd}{PDCD}{Primal-dual coordinate descent}{-}

\myacronym{rv}{RV}{random variable}{random variables}
\myacronym{qr}{QR}{quantile regression}{-}
\myacronym{iid}{\emph{iid}}{independent and identically distributed}{-}
\myacronym{cadlag}{càdlàg}{right continuous with left limits}{}
\myacronym{pdf}{pdf}{probability density function}{probability density functions}
\myacronym{cdf}{cdf}{cumulative distribution function}{cumulative distribution functions}
\myacronym{mle}{MLE}{maximum likelihood estimator}{maximum likelihood estimators}
\myacronym{emalg}{EM}{expectation-maximization algorithm}{-}

\myacronym{psd}{PSD}{positive semi-definite}{-}
\myacronym{pd}{PD}{positive definite}{i}

\usepackage[margin=3cm]{geometry}

\usepackage{authblk}

\title{Proximal boosting: aggregating weak learners to minimize non-differentiable losses}
\author{Erwan Fouillen}
\author{Claire Boyer}
\author{Maxime Sangnier}
\affil{Université de Paris and Sorbonne Université, CNRS, Laboratoire de Probabilités, Statistique et Modélisation, F-75013 Paris, France}

\begin{document}
	\maketitle

	\begin{abstract}
Gradient boosting is a prediction method that iteratively combines weak learners to produce a complex and accurate model.
From an optimization point of view, the learning procedure of gradient boosting mimics a gradient descent on a functional variable.
This paper proposes to build upon the proximal point algorithm, when the empirical risk to minimize is not differentiable, in order to introduce a novel boosting approach, called \emph{proximal boosting}.
It comes with a companion algorithm inspired by \citep{grubb_generalized_2011} and called \emph{residual proximal boosting}, which is aimed at better controlling the approximation error.
Theoretical convergence is proved for these two procedures under different hypotheses on the empirical risk and advantages of leveraging proximal methods for boosting are illustrated by numerical experiments on simulated and real-world data.
In particular, we exhibit a favorable comparison over gradient boosting regarding convergence rate and prediction accuracy.


	\end{abstract}

	\section{Introduction}
Boosting is a celebrated machine learning technique, both in statistics and data science.
In broad outline, boosting \neurocomp{sequentially} combines simple models (called weak learners) to build a more complex and accurate model.
This assembly is performed iteratively, taking into account the performance of the model built at the previous iteration.
The way this information is considered leads to several variants of boosting, the most famous of them being Adaboost \citep{freund_decision-theoretic_1997} and gradient boosting \citep{friedman_greedy_2001}.

The reason of the success of boosting is twofold:
i) from the statistical point of view, boosting is an additive model with an iteratively growing complexity.
It is thus possible to reduce the bias of the risk while controlling its variance.
This is a noticeable advantage over very complex models such as nonparametric methods.
ii) from the data science perspective, fitting a boosting model is computationally cheap, making it possible to be used on large datasets.
In contrast, it can quickly achieve sufficiently complex models to be able to perform accurately on difficult learning task.
As an ultimate feature, the iterative process makes finding the frontier between under and overfitting quite easy.
In particular, gradient boosting combined with decision trees (often referred to as gradient tree boosting) is currently regarded as one of the best off-the-shelf learning techniques 
\neurocomp{for tabular data in several real-world situations ranging from data challenges to tangible applications in urbanization \citep{ikeagwuani_resilient_2021,rajendran_predicting_2021}, renewable energy \citep{tyralis_boosting_2021,chen_output_2022} and medical care \citep{ahamad_machine_2020,awal_novel_2021,santana_classification_2021}.}

As explained by \citet{biau_accelerated_2018}, gradient boosting has its roots in Freund and Schapire's work on combining classifiers, which resulted in the Adaboost algorithm \citep{freund_decision-theoretic_1997,schapire_strength_1990,freund_boosting_1995,freund_experiments_1996}.
Later, Friedman and colleagues developed a novel boosting procedure inspired by the numerical optimization literature, and nicknamed \emph{gradient boosting} \citep{friedman_greedy_2001,friedman_additive_2000,friedman_stochastic_2002}.
Such a connection of boosting to statistics and optimization was already stated in several previous analyses by Breiman \citep{breiman_arcing_1997,breiman_arcing_1998,breiman_prediction_1999,breiman_infinite_2000,breiman_population_2004} and reviewed as functional optimization \citep{mason_functional_2000,mason_boosting_2000,meir_introduction_2003,buhlmann_boosting_2007}:
boosting can be seen as an optimization procedure (similar to gradient descent), aimed at minimizing an empirical risk over the set of linear combinations of weak learners.
In this respect, a few theoretical studies prove the convergence, from an optimization point of view, of boosting procedures \citep{zhang_general_2002,zhang_sequential_2003,wang_functional_2015}
and particularly of gradient boosting \citep{temlyakov_greedy_2014,biau_optimization_2021}.
Let us remark that rates of convergence of gradient boosting are known for smooth and strongly convex risks \citep{grubb_generalized_2011,ratsch_convergence_2002}.

\neurocomp{Since the invention of gradient boosting, several variants have been emerging (see for instance \citep{buhlmann_boosting_2003,zhang_boosting_2005,gao_multiclass_2011,wang_rescaled_2019} to cite only a few) up to very recent studies concerning large-scale regression with boosted histograms \citep{cai_boosted_2020,cui_gbht_2021,hang_gradient_2021}.
The statistical properties of boosting algorithms have been addressed many times (for instance in \citep{buhlmann_boosting_2003,park_l_2_2009,lin_boosted_2019}) and are still under consideration as a modern topic of statistical learning \citep{cai_boosted_2020,cui_gbht_2021,hang_gradient_2021,zeng_fully_2022}.}

In practice, the number of weak learners used in gradient boosting (and variants) controls the statistical complexity of the final predictor but also the number of optimization steps performed in order to minimize the empirical risk.
While controlling the latter is a natural way to regularize the method and to enhance its generalization properties,
tuning the former makes it possible to stop the optimization algorithm before convergence, which is known in many areas as early stopping.
This technique can be seen as an iterative regularization mechanism also used to prevent overfitting \citep{lin_iterative_2016}.
As a consequence, besides its approximation capability, the statistical performance of gradient boosting deeply relies on the algorithm employed.

That being said, one may wonder if gradient descent is really a good option.
Following this direction, several alternatives have been proposed, such as replacing gradient descent by the Frank-Wolfe algorithm \citep{wang_functional_2015}, incorporating second order information \citep{chen_xgboost:_2016}, and applying Nesterov's acceleration \citep{biau_accelerated_2018,lu_accelerating_2020}.
While all these variants rely on differentiable loss functions, \citet{grubb_generalized_2011} discusses the limitations of boosting with gradient descent in the non-differentiable setting, and tackle these issues by proposing two modified versions of (sub)gradient boosting, consisting in reprojecting the error made when approximating the subgradients by weak learners.
The contribution of the work described here is to go a step forward
by proposing novel procedures to efficiently learn boosted models with non-differentiable loss functions.

To go into details, Section~\ref{sec:problem} reviews boosting with respect to the empirical risk minimization principle and illustrates the flaw of the current learning procedure in a simple non-differentiable case: least absolute deviations.
\neurocomp{Building upon a background on non-smooth optimization,  Section~\ref{sec:algorithm} encloses the main contribution of this paper:}
adapting the proximal point algorithm \citep{nesterov_introductory_2004}
to boosting.
The proposed method is nicknamed \emph{proximal boosting} and comes with a variant, called \emph{residual proximal boosting}, inspired by \citet{grubb_generalized_2011}.
A second contribution is to prove convergence rates (from an optimization perspective) of proximal and residual proximal boosting under different hypotheses on the loss function (see Section~\ref{sec:cv}).
Finally, the numerical study described in Section~\ref{sec:numerical} shines a light on advantages and limitations of the proposed boosting procedures.
\new{As a by-product, we also consider adapting Nesterov's acceleration to proximal boosting, such as in accelerated gradient boosting \citep{biau_accelerated_2018}.
Even though our proposed algorithm performs better than that of \citep{biau_accelerated_2018}, we observe divergence on the training set
(as this is the case for accelerated gradient boosting \citep{biau_accelerated_2018,lu_accelerating_2020})
and no particular gain in accuracy.}


	\section{Problem and notation}
		\label{sec:problem}
Let $\X$ be an arbitrary input space and $\Y \subseteq \R$ an output space.
Given a pair of random variables $(X, Y) \in \X \times \Y$, supervised learning aims at explaining $Y$ given $X$, thanks to a measurable function $f_0 \colon \X \to \R$.
In this context, $f_0(X)$ may represent several quantities, depending on the task at hand, for which the most notable examples are the conditional expectation $x \in \X \mapsto \expe[Y|X=x]$ and the conditional quantiles of $Y$ given $X$ for regression, as well as the regression function $x \in \X \mapsto \prob(Y=1|X=x)$ for $\pm 1$-classification.
Often, this target function $f_0$ is a minimizer of the risk $\expe(\ell(Y, f(X)))$ over all measurable functions $f$, where $\ell: \R \times \R \to \R$ is a suitable convex loss function (respectively the square function and the pinball loss in the regression examples previously mentioned).

Since the distribution of $(X,Y)$ is generally unknown, the minimization of the risk is out of reach. One would rather deal with its empirical version instead.
Let $\{(X_i, Y_i)\}_{1 \le i \le n} \subseteq \X \times \Y$ be a training sample of pairs $(X_i, Y_i)$ independent and identically distributed according to the distribution of $(X, Y)$, \(\fset\) the set of functions from \(\X\) to \(\R\) and $\F \subseteq \fset$ a class of functions.
In this work, we consider estimating $f_0$ \replace{thanks to}{by means of} an additive model $f^\star$ (that is $f^\star = \sum_{t=0}^T w_t g_t$, where $T$ is an unknown integer and $(w_t, g_t)_t \subseteq \R \times \F$ is an unknown sequence of weights and weak learners) by solving the following optimization problem:
\begin{opb}{opb:general}
  \minimize{f \in \lin \F} & C(f),
\end{opb}
where
\[
  C(f)
  = \frac 1n \sum_{i=1}^n \ell(Y_i, f(X_i))
\]
is the empirical risk and
\[
  \lin \F = \left\{ \sum_{t=1}^m w_t g_t : w \in \R^m, (g_1, \dots, g_m) \in \F^m, m \in \N \right\}
\]
is the set of all linear combinations of functions in $\F$ ($\N$ being the set of non-negative integers).

\begin{figure}
  \center
  \includegraphics[width=0.8\textwidth]{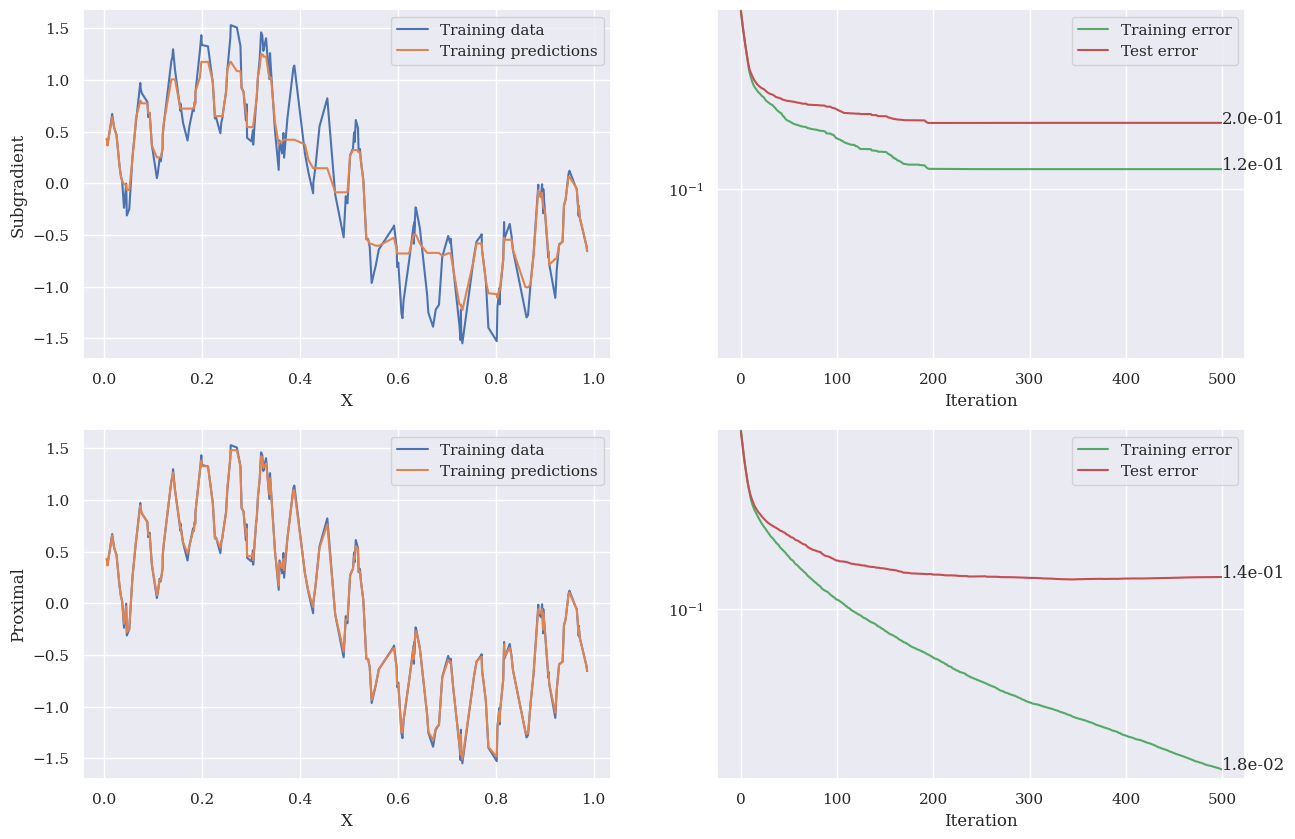}
  \caption{Predicted values and training error of a boosting machine trained with a subgradient (top) and a proximal-based method (bottom).}
  \label{fig:intro}
\end{figure}

As a simple example, let us consider the regression model $Y = \sin(2\pi X) + \sin(32\pi X) + \epsilon$, where $X$ is uniformly distributed on $[0, 1]$ and $\epsilon$ is normally distributed and independent of $X$.
We aim at solving:
\begin{opb*}
  \minimize{f \in \lin \F}
  & \frac 1n \sum_{i=1}^n |Y_i - f(X_i)|,
\end{opb*}
with $\F$ being the set of regression trees of depth less than $3$.

Two boosting machines $f_T = \sum_{t=0}^T w_t g_t$ are learned (with $T$ fixed to $500$): a traditional one with a subgradient-type method (Algorithm~\ref{alg:gradient}, \neurocomp{Section~\ref{sec:gradient_boosting}}), and another with the proposed proximal-based procedure (Algorithm~\ref{alg:prox}, \neurocomp{Section~\ref{sec:proposition}}).
Figure~\ref{fig:intro} depicts the prediction of $f_T$ (left) and the training error $C(f_t) = \frac 1n \sum_{i=1}^n |Y_i - f_t(X_i)|$ along the iterations $t$ (right, green curve).

From an optimization perspective, it appears clearly that the subgradient method fails to minimize the empirical risk (prediction is far from the data and the training error attains a plateau at \(1.2 \cdot 10^{-1}\))
while the proximal-based procedure constantly improves the objective.
The subgradient method faces a flaw in convergence, in all likelihood due to non-differentiability of the absolute function $|\cdot|$.
This simple example illustrates, inside the boosting paradigm, a well-known fact in numerical optimization: proximal-based algorithms prevail over subgradient techniques for non-differentiable objective functions.

Beyond optimization, proximal boosting also outperforms gradient boosting from a statistical perspective since it achieves a lower test error (red curve in the right side of Figure~\ref{fig:intro}).

	\section{Algorithms}
		\label{sec:algorithm}
There is an ambiguity in \ref{opb:general}, since it is a functional optimization problem but, in practice, we do not necessarily have the mathematical tools to apply standard optimization procedures (in particular concerning differentiation of $C$).
For this reason, $C$ is often regarded as a function from $\R^n$ to $\R$, considering that it depends on $f$ only through the vector $f(X_1^n) = (f(X_1), \dots, f(X_n)) \in \R^n$. 
To make this remark more precise, let, for all $z \in \R^n$, $D(z) = \frac 1n \sum_{i=1}^n \ell(Y_i, z_i)$.
Then,
for any $f \in \fset$, $C(f) = D(f(X_1^n))$. 

Having this remark in mind helps solving \ref{opb:general}, for instance considering that taking the gradient of $C$ with respect to $f$ is roughly equivalent to differentiating $C$ with respect to $f(x)$
(for all observed $x \in \{X_1, \dots, X_n\}$),
thus taking in fact the usual gradient of $D$.
Doing so, the only requirement is to match the vectors appearing in standard optimization procedures with functions from \(\fset\).
In particular, given a vectorial gradient $\nabla D(f(X_1^n))$ ($f \in \fset$), one has to find a function $g \in \fset$ that correctly represents it, \ie such that $g(X_1^n) \approx \nabla D(f(X_1^n))$.
This principle is at the heart of functional optimization methods such that the ones used in boosting \citep{mason_functional_2000}.

From now on, all necessary computations of $C$ with respect to $f$ can be forwarded to $D$.
For instance, if $\ell$ is differentiable with respect to its second argument, we can define, for all $f \in \fset$, the functional gradient of $C$ as $\fgrad C(f) = \nabla D(f(X_1^n))$.
On the contrary, if $\ell$ is not differentiable, we may consider a subgradient of $C$ at $f$, denoted $\fsub C(f)$ and defined as any subgradient of $D$ at $f(X_1^n)$.

In the forthcoming sections, a common first order optimization algorithm is reviewed.
Then, it is explained how to build different procedures for solving \ref{opb:general}, according to the properties of the loss function $\ell$.
\neurocomp{For the sake of readability, the algorithms introduced in the next sections are recapped in Table~\ref{tab:recap}.}

\begin{table}
  \begin{tabular}{cp{.8\textwidth}}
    \textsc{Algorithm}  & \textsc{Description} \\ \hline
    \ref{alg:gradient} & Gradient boosting with convergence rate \(O(1/t)\) for smooth losses \\
    \ref{alg:prox} & Proximal boosting with convergence rate \(O(1/t)\) for smooth losses \\
    \ref{alg:prox_res} & Proximal boosting with convergence rate \(O(1/\sqrt t)\) for non-smooth losses \\
    \ref{alg:prox_acc} & Proximal boosting with Nesterov's acceleration \\
    \ref{alg:prox_meta} & Numerical implementation of Algorithms~\ref{alg:gradient}, \ref{alg:prox} and \ref{alg:prox_res} with shrunk step size \\
    \ref{alg:app_prox} & Abstract algorithm for the proof of Theorem~\ref{thm:cv_results} \\
    \ref{alg:app_prox_acc} & Abstract algorithm for the proof of Theorem~\ref{thm:cv_results2} \\
    \ref{alg:prox_acc_practice} & Numerical implementation of Algorithm~\ref{alg:prox_acc} with shrunk step size \\
  \end{tabular}
  \caption{Summary of algorithms.}
  \label{tab:recap}
\end{table}

\subsection{The proximal gradient method}
  \label{sec:optim}
  Let us assume for a while that we want to minimize the function $g + h$, where $g \colon \R^d \to \R$ is convex and differentiable (with $L$-Lipschitz continuous gradient, $L>0$), and $h \colon \R^d \to \R\cup\{+\infty\}$ is convex and lower semi-continuous.
  \replace{}{Besides, let us define the proximal operator of $h$ by:
  \[
    \prox_{h} (x) = \argmin_{u \in \R^d} \left\{ h(u) + \frac 12 \norm{u - x}^2\right\},
    \qquad \forall x \in \R^d,
  \]
  \neurocomp{where \(\norm{\cdot}\) is the Euclidean norm.
  This operator }
  is well defined by convexity and lower semi-continuity of $h$ \citep{combettes_signal_2005}.
  }
  Then, the iterative procedure defined by choosing any
  $x_0 \in \R^d$ and by setting for all $t \in \N$:
  \[
    x_{t+1} = \prox_{\gamma_{t+1} h}(x_t - \gamma_{t+1} \nabla g(x_t)),
  \]
  where $\gamma_{t+1} \in (0, 2/L)$,
  is known as the proximal gradient method, and converges to a minimizer of $g+h$ in $O(1/t)$ \citep{nesterov_introductory_2004}.
  More formally, assuming that $g+h$ has a minimizer $x^\star$, then $(g+h)(x_t)-(g+h)(x^\star) = O(1/t)$.

  Depending on the properties of the objective function to minimize, the procedure described before leads to two simple algorithms:
  \begin{itemize}
    \item the gradient method ($h=0$):
    \[
      x_{t+1} = x_t - \gamma_{t+1} \nabla g(x_t),
    \]
    minimizes a single function $g$ as soon as it is convex and differentiable with Lipschitz continuous gradient;
    \item the proximal point algorithm ($g=0$):
    \begin{equation}
      x_{t+1} = \prox_{\gamma_{t+1} h}(x_t) = x_t - \gamma_{t+1} \left[ \frac{1}{\gamma_{t+1}} \left( x_t - \prox_{\gamma_{t+1} h}(x_t) \right) \right],
      \label{equ:proximal_iteration}
    \end{equation}
    minimizes a single function $h$, which is only required to be convex and lower semi-continuous (in this case, there is no restriction on the step size $\gamma_{t+1}$, except being positive).
  \end{itemize}

  The proximal gradient method (as well as its two special cases) has the asset to be a descent method: at each iteration, the objective function monotonically decreases, meaning that $(g+h)(x_{t+1}) \le (g+h)(x_t)$, with convergence rate $O(1/t)$.
  In particular, this is true when minimizing a single convex and lower semi-continuous function $h \colon \R^d \to \R$, even if it is not differentiable, with the iteration given in Equation~\eqref{equ:proximal_iteration}.

  This has to be put in contrast with the subgradient method:
  \begin{equation}
    x_{t+1} = x_t - \gamma_{t+1} \widetilde \nabla h (x_t),
    \label{equ:subgradient_iteration}
  \end{equation}
  where $\gamma_{t+1} > 0$ and $\widetilde \nabla h (x_t)$ is any subgradient of $h$ at $x_t$.
  This procedure, which is very similar to the gradient descent but replacing the gradient by any subgradient, has a convergence rate $O(1/\sqrt t)$ in the best case \citep{nesterov_introductory_2004}.
  In addition,
  this rate is
  tight for this optimization procedure:
  it cannot be improved without extra assumptions on $h$ \citep[Theorem~3.2.1]{nesterov_introductory_2004}.

  This remark motivates the use of procedures different from the subgradient method when minimizing a non-differentiable function $h$, such as the proximal point algorithm (described in Equation~\eqref{equ:proximal_iteration}).
  This motivation is emphasized by the fact that moving from the subgradient to the proximal point method
  only requires to replace the update direction $\widetilde \nabla h(x_t)$ by $\frac{1}{\gamma_{t+1}} (x_t - \prox_{\gamma_{t+1} h}(x_t))$.
  This observation is the cornerstone of the boosting algorithms proposed in Section~\ref{sec:proposition}.

\subsection{Gradient boosting}
  \label{sec:gradient_boosting}
  Let $\F_0$ be the set of constant functions on $\X$ and assume that $\F_0 \subseteq \F$.
  Then, a simple procedure to approximately solve \ref{opb:general} is gradient boosting, described in Algorithm~\ref{alg:gradient} \citep{friedman_greedy_2001,mason_boosting_2000}.
  It builds the requested additive model in an iterative fashion, by imitating a gradient method (or subgradient method if $\ell$ is not differentiable with respect to its second argument).
  At each iteration $t$, Algorithm~\ref{alg:gradient} finds a function $g_{t+1}$ that approximates the opposite of a subgradient of $C$ (also called pseudo-\replace{residues}{residuals}) and adds it to the model $f_t$ with a
  positive weight $w_{t+1} = \gamma_{t+1}$.
  At the end of the procedure, the proposed estimator of $f_0$ is $f_T = \sum_{t=0}^T w_t g_t$, with $w_0 = 1$.


\begin{algorithm}[ht]
  \begin{algorithmic}[1]
    \REQUIRE
    \(\gamma_1, \dots, \gamma_T > 0\) (gradient steps).
    \STATE Set $f_0 \in \argmin_{g \in \F_0} C(g)$
    \COMMENT{initialization}.
    \FOR{$t=0$ \TO $T-1$}
      \STATE Compute $r \gets -\fsub C (f_t)$
      \COMMENT{pseudo-residuals}.
      \STATE Compute \(g_{t+1} \in \argmin_{g \in \F} \norm{g(X_1^n) \replace{+ \fsub C (f_t)}{-r}}\).
      \STATE Set $f_{t+1} \gets f_t + \gamma_{t+1} g_{t+1}$.
      \COMMENT{update}.
    \ENDFOR
    \ENSURE $f_T$.
  \end{algorithmic}
  \caption{Gradient boosting.}
  \label{alg:gradient}
\end{algorithm}

  There are several manners to schedule the gradient steps $\gamma_{t+1}$, including being adaptively fixed thanks to a line search.
  This is discussed in
  Section~\ref{sec:numerical}.

\subsection{Boosting with non-differentiable loss functions}
  \label{sec:proposition}
  When the function $\ell$ is not differentiable with respect to its second argument, gradient boosting just uses a subgradient $\fsub C(f_t)$ instead of the gradient $\fgrad C(f_t)$.
  This is, of course, convenient but as explained previously, far from leading to interesting convergence behaviors in practice.
  For this reason, we propose a new procedure for non-differentiable loss functions $\ell$, which consists in adapting the proximal point algorithm \citep{nesterov_introductory_2004} to functional optimization.

  For any $f \in \fset$, let $\proxdir{\lambda} C (f) = \frac{1}{\lambda} \left( f(X_1^n) - \prox_{\lambda D}(f(X_1^n)) \right)$, where $\lambda > 0$ is a parameter.
  The simple idea underlying the proposed algorithm, nicknamed \emph{proximal boosting},
  \replace{is to replace $\fsub C(f_t)$ by $\proxdir{\lambda_{t+1}} C(f_t)$, remarking that
  $f_t(X_1^n) - \lambda_{t+1} \proxdir{\lambda_{t+1}} C(f_t) = \prox_{\lambda_{t+1} D}(f_t(X_1^n))$ is exactly the iteration update of the proximal point method.}{is that the only difference between subgradient and proximal point methods is the update direction of the optimization variable, which is respectively $\fsub C(f_t)$ or $\proxdir{\lambda_{t+1}} C (f_t)$, where \(\lambda_{t+1} > 0\) is a proximal step.
  Thus, proximal boosting computes the pseudo-residuals based on $\proxdir{\lambda_{t+1}} C (f_t)$ instead of $\fsub C(f_t)$
  and leaves the rest unchanged, as described in Algorithm~\ref{alg:prox}.}


\begin{algorithm}[ht]
  \begin{algorithmic}[1]
    \REQUIRE
    $\lambda_1, \dots, \lambda_T > 0$ (proximal steps).
    \STATE Set $f_0 \in \argmin_{g \in \F_0} C(g)$
    \COMMENT{initialization}.
    \FOR{$t=0$ \TO $T-1$}
      \STATE Compute $r \gets -\proxdir{\lambda_{t+1}} C(f_t)$
      \COMMENT{pseudo-\replace{residues}{residuals}}.
      \STATE Compute \(g_{t+1} \in \argmin_{g \in \F} \norm{g(X_1^n) - r}\).
      \STATE Set $f_{t+1} \gets f_t + \lambda_{t+1} g_{t+1}$.
    \ENDFOR
    \ENSURE $f_T$.
  \end{algorithmic}
  \caption{Proximal boosting.}
  \label{alg:prox}
\end{algorithm}

  \new{
  While Algorithm~\ref{alg:prox} is very intuitive and proved to converge at the expected rate for differentiable loss functions (see Section~\ref{sec:cv}), a rate of convergence cannot be exhibited for non-differentiable loss functions.
  To remedy this limitation, we now introduce a variant of Algorithm~\ref{alg:prox}, named \emph{residual proximal boosting} (see Algorithm~\ref{alg:prox_res}) and inspired by \citep{grubb_generalized_2011}, which incorporates a mechanism making it possible to control the approximation error made at each iteration and to obtain a convergence rate under weak assumptions (see Section~\ref{sec:cv}).
  }
  In practice, it consists in augmenting the pseudo-residuals with the approximation error \(\Delta_t\) of the previous iteration, also called residual.


\begin{algorithm}[ht]
  \begin{algorithmic}[1]
    \REQUIRE
    $\lambda_1, \dots, \lambda_T > 0$ (proximal steps).
    \STATE Set $f_0 \in \argmin_{g \in \F_0} C(g)$,
    \(\Delta_0 \gets 0\)
    \COMMENT{initialization}.
    \FOR{$t=0$ \TO $T-1$}
      \STATE Compute $r \gets -\proxdir{\lambda_{t+1}} C(f_t)$
      \COMMENT{pseudo-\replace{residues}{residuals}}.
      \STATE Compute \(g_{t+1} \in \argmin_{g \in \F} \norm{g(X_1^n) - (r + \Delta_t)}\).
      \STATE Set $f_{t+1} \gets f_t + \lambda_{t+1} g_{t+1}$.
      \STATE Set \(\Delta_{t+1} \gets r + \Delta_t - g_{t+1}(X_1^n)\).
    \ENDFOR
    \ENSURE $f_T$.
  \end{algorithmic}
  \caption{Residual proximal boosting.}
  \label{alg:prox_res}
\end{algorithm}

  \new{As a by-product and along the same line as accelerated gradient boosting \citep{biau_accelerated_2018}, we remark that this is possible to incorporate Nesterov's acceleration \citep{nesterov_method_1983, beck2009fast} to proximal boosting, in order to speed up the convergence and to aggregate less weak learners.
  In practice, Algorithm~\ref{alg:prox_acc} is similar to Algorithm~\ref{alg:prox} but computes the proximal step at the auxiliary function \(h_t\) instead of \(f_t\).
  \(h_{t+1}\) is then obtained by a momentum tuned by the coefficient \(\alpha_t\), defined recursively by
  \begin{equation}
    \begin{cases}
      \beta_0 = 0 \\
      \beta_{t+1} = \frac{1 + \sqrt{1 + 4 \beta_t^2}}{2}, t \in \N \\
      \alpha_{t+1} = \frac{\beta_t - 1}{\beta_{t+1}}, t \in \N.
    \end{cases}
    \label{equ:nesterov}
  \end{equation}
  Algorithm~\ref{alg:prox_acc} returns an estimator $f_T = \sum_{t=0}^T w_t g_t$ where the weights $w_0, \dots, w_T$ are now given by a recursive formula (see Appendix~\ref{app:proof_weights}).}

  The convergence rate of the accelerated proximal point method is $O(1/t^2)$, which prevails over that of the vanilla version of the proximal point method from an optimization point of view.
  However, as it will be observed in Section~\ref{sec:numerical}, the boosting procedure proposed in Algorithm~\ref{alg:prox_acc} inherits the same drawbacks as accelerated gradient boosting and does not seem reliable.
  Importantly, it is prone to divergence.


\begin{algorithm}[ht]
  \begin{algorithmic}[1]
    \REQUIRE
    $\lambda_1, \dots, \lambda_T > 0$ (proximal steps).
    \STATE Set $f_0 = h_0 \in \argmin_{g \in \F_0} C(g)$
    \COMMENT{initialization}.
    \FOR{$t=0$ \TO $T-1$}
      \STATE Compute $r \gets -\proxdir{\lambda_{t+1}} C(h_t)$
      \COMMENT{pseudo-\replace{residues}{residuals}}.
      \STATE Compute \(g_{t+1} \in \argmin_{g \in \F} \norm{g(X_1^n) - r}\).
      \STATE Set $f_{t+1} \gets h_t + \lambda_{t+1} g_{t+1}$. \label{lin:update3}
      \STATE Set $h_{t+1} \gets f_{t+1} + \alpha_{t+1}(f_{t+1} - f_{t})$.
    \ENDFOR
    \ENSURE $f_T$.
  \end{algorithmic}
  \caption{Accelerated proximal boosting.}
  \label{alg:prox_acc}
\end{algorithm}


	\section{Convergence results}
		\label{sec:cv}
This section is dedicated to the theoretical convergence of the two proposed algorithms: proximal boosting (Algorithm~\ref{alg:prox}) and residual proximal boosting (Algorithm~\ref{alg:prox_res}).

A preliminary result on the convergence of the proximal boosting technique can be easily derived upon previous work by \citet{rockafellar1976monotone}: it requires to control the error introduced by considering an approximated direction of optimization instead of the true proximal step, and could be stated as follows in the case of Algorithm~\ref{alg:prox}.

\begin{theorem}[{\cite[Theorem 1]{rockafellar1976monotone}}]
	Let $(f_t)_{t}$ be any sequence generated by Algorithm~\ref{alg:prox}
	and define for any iteration $t$:
	\[
		\varepsilon_{t+1} = \norm{g_{t+1}(X_1^n) + \proxdir{\lambda_{t+1}} C(f_t)}.
	\]
	Suppose that $(f_t(X_1^n))_t$ is bounded and that
	\begin{align}
	\label{equ:cond_cv_rocka1}
	\sum_{t=0}^{+\infty} \varepsilon_t < + \infty.
	\end{align}
	Then,
	$$
	\lim_{t\to \infty} C(f_t) = \inf_{f \in \lin \F} C(f).
	$$
	\label{thm:rockafellar_weak}
\end{theorem}

Theorem~\ref{thm:rockafellar_weak} states that as soon as the approximation errors $(\varepsilon_t)_t$ converge to $0$ quicker than $1/t$, then the sequence $(C(f_t))_t$ converges to a minimum of $C$.
However, with a better control of the approximation errors $(\varepsilon_t)_t$, a rate of convergence can be derived for Algorithm~\ref{alg:prox}.
This is the role of the following assumption, which is common in the boosting literature to characterize the approximation capacity of the class \(\F\) \citep{grubb_generalized_2011}.
\begin{assumptions}
	\hyp{A} There exists $\zeta \in (0, 1]$ such that:
	 \[
	 	\forall r \in \R^n,
		\qquad
		\exists g \in \F : \norm{g(X_1^n) - r}^2 \le (1-\zeta^2) \norm{r}^2.
	 \]
	  \label{hyp:edge0}
\end{assumptions}
A set of weak learners \(\F\) satisfying Assumption~\ref{hyp:edge0} is said to have edge \(\zeta\).

Now, we provide a convergence result for Algorithm~\ref{alg:prox}, based on smoothness properties:
a functional $C$ of the form $C(f) = D(f(X_1^n))$, for all $f \in \fset$, is said $L$-smooth (for some $L>0$) if \(D\) is differentiable and for all $x, x' \in \R^n$,
\[
	D(x') \le D(x) + \iprod{\nabla D(x)}{x'-x} + \frac{L}{2} \norm{x'-x}^2,
\]
and $\kappa$-strongly convex (for some $\kappa>0$) if
\[
	D(x') \ge D(x) + \iprod{\nabla D(x)}{x'-x} + \frac{\kappa}{2} \norm{x'-x}^2,
\]
\neurocomp{where \(\iprod{\cdot}{\cdot}\) refers to the inner product.}
The convergence rate stated hereafter is based on an original result presented and proved in Appendix~\ref{app:proof_cv}.

\begin{theorem}\label{thm:cv_results}
	Assume that \ref{hyp:edge0} is granted, $C$ is $L$-smooth and $\kappa$-strongly convex for some $L>0$ and $\kappa>0$.
	Let $(f_t)_{t}$ be any sequence generated by Algorithm~\ref{alg:prox}
	and assume that there exists $f^\star \in \argmin_{f \in \lin \F} C(f)$.
	Then, choosing $\lambda_t = \frac{\zeta^2}{8L}$ leads to:
	\[
		C(f_T) - C(f^\star) \leq
		\left( 1- \frac{\zeta^4  \kappa}{21L} \right)^T
		\left( C(f_0) - C(f^\star) \right).
	\]
\end{theorem}
\begin{proof}
	Given that $\forall f \in \fset: C(f) = D(f(X_1^n))$
	\neurocomp{and Assumptions~\ref{hyp:edge}, \ref{hyp:str_smooth} and \ref{hyp:str_cvx} are granted for \(D\) (respectively by Assumption~\ref{hyp:edge0}, $L$-smoothness and $\kappa$-strong convexity of \(C\)),}
	this is an application of Theorem~\ref{thm:cv_app_prox_sc_sm} (see Appendix~\ref{app:proof_cv}) to the function $D$.
\end{proof}

Theorem~\ref{thm:cv_results} states that proximal boosting has a linear convergence rate under smoothness and strong convexity assumptions.
This result was expected since gradient boosting has the same convergence rate under these assumptions \citep{grubb_generalized_2011}.

Admittedly, these two assumptions are restrictive for an algorithm designed for non-differentiable loss functions.
However, our analysis revealed that they seem necessary to control the impact of the approximation error on the convergence.
Consequently, proving convergence for proximal boosting under weaker assumptions on the objective function \(C\) (see thereafter) requires to modify Algorithm~\ref{alg:prox}.
This is the role of Algorithm~\ref{alg:prox_res}, as introduced in Section~\ref{sec:algorithm}.

A functional $C$ of the form $C(f) = D(f(X_1^n))$, for all $f \in \fset$, is said to be \(G\)-Lipschitz continuous (for some \(G>0\)) if for all $x, x' \in \R^n$,
\[
	|D(x) - D(x')| \le G \norm{x - x'}.
\]
A convergence rate for residual proximal boosting (Algorithm~\ref{alg:prox_res}) can be derived from this weak property, as stated in Theorem~\ref{thm:cv_results2} (which is based on an original result presented and proved in Appendix~\ref{app:proof_cv}).

\begin{theorem}\label{thm:cv_results2}
	Assume that \ref{hyp:edge0} is granted, $C$ is convex and \(G\)-Lipschitz continuous for some \(G>0\).
	Let $(f_t)_{t}$ be any sequence generated by Algorithm~\ref{alg:prox_res}
	and \(f_{\text{best}} \in \argmin_{1 \le t \le T} C(f_t)\).
	Assume that there exists $f^\star \in \argmin_{f \in \lin \F} C(f)$ and
	that \(\norm{f_t(X_1^n)} \le  R\) and \(\norm{f^\star(X_1^n)} \le R\) for some \(R>0\) and all \(t\).
	Then, choosing $\lambda_t = \frac{1}{\sqrt{t}}$ leads to:
	\[
		C(f_{\text{best}}) - C(f^\star) \leq
		\frac{2R^2}{\sqrt T} + \frac{40G^2}{\zeta^4 \sqrt T} + \frac{2G^2}{\zeta^4 T^{\frac 32}}.
	\]
\end{theorem}
\begin{proof}
	Given that $\forall f \in \fset: C(f) = D(f(X_1^n))$
	\neurocomp{and Assumptions~\ref{hyp:edge} and \ref{hyp:lipf} are granted for \(D\) (respectively by Assumption~\ref{hyp:edge0} and $G$-Lipschitz continuity of \(C\)),}
	this is an application of Theorem~\ref{thm:cv_app_prox_l} (see Appendix~\ref{app:proof_cv}) to the function $D$.
\end{proof}

Theorem~\ref{thm:cv_results2} states that the best aggregation returned by residual proximal boosting has sublinear convergence rate (more precisely \(O(1/\sqrt t)\)) under Lipschitz continuity assumption.
On the one hand, this rate is similar to that of residual gradient boosting \citep{grubb_generalized_2011}, showing that our approach is theoretically competitive with the state-of-the art regarding boosting with non-differentiable cost functions.
On the other hand, this result
is quite pessimistic regarding the empirical performance of Algorithm~\ref{alg:prox_res}:
Section~\ref{sec:numerical} will show that, in practice, linear convergence (as stated by Theorem~\ref{thm:cv_results}) is often observed numerically, even though the loss function is not differentiable.
This is perfectly consistent with our initial intuition: boosting better handles the non-differentiability of the objective function by using the proximal operator instead of any subgradient.

\begin{remark}
\neurocomp{
    Since the convergence rate of the proximal point method for non-smooth functions is \(O(1/t)\) (respectively \(O(1/ \sqrt t)\) for the subgradient method), one may expect that proximal boosting converges in \(O(1/t)\) (while subgradient boosting is in \(O(1/\sqrt t)\) \citep{grubb_generalized_2011})
but the previous result states a worst case convergence rate in \(O(1/\sqrt t)\).
}

\neurocomp{
    The latter is in fact not that surprising: for \(L\)-smooth and \(\kappa\)-strongly convex objectives, gradient descent converges in \(O\left(\left( 1 - \frac{\kappa}{L} \right)^t \right)\) while gradient boosting converges in \(O(1/t)\).
    This highlights that the approximation step (represented by the operator \(P\) below) used in boosting iterations is prone to damage the convergence rate.
}

\neurocomp{
    More formally, consider an objective function \(f\) and two iterations \(x_{t+1}^{theo}=x_t - \gamma_t d_t\) and \(x_{t+1}=x_t - \gamma_t P(d_t)\),  where \(P\) is an approximation operator.
    The rate in \(O(1/t)\) for gradient descent and the proximal point method is linked to the capability
    to control the error \(\varepsilon(\tilde \nabla f(x_{t+1}^{theo}), d_t)\) between a subgradient at \(x_{t+1}^{theo}\), denoted \(\tilde \nabla f(x_{t+1}^{theo})\), and the direction of descent at \(x_t\), denoted \(d_t\).
}

\neurocomp{
    The error \(\varepsilon(\tilde \nabla f(x_{t+1}^{theo}), d_t)\) is (i) controlled under the assumption of Lipschitz continuous gradients in the case of gradient descent; (ii) equal to 0,
    (\(\varepsilon(\tilde \nabla f(x_{t+1}^{theo}), d_t)=0\)) for the proximal point method (the proximal direction of descent is exactly a subgradient at \(x_{t+1}^{theo}\)).
    If this error cannot be controlled tightly, we may end up with a \(O(1/\sqrt t)\) convergence rate.
    This is exactly the case for the subgradient method,
    for which \(\varepsilon(\tilde \nabla f(x_{t+1}^{theo}), d_t)=\varepsilon(\tilde \nabla f(x_{t+1}^{theo}), \tilde \nabla f(x_{t}))\) (which is a difference between two subgradients) is only bounded by a constant.
}

\neurocomp{
    In proximal and subgradient boosting (second iteration as defined above), this error can be decomposed into three parts:
    \[
      \varepsilon(\tilde \nabla f(x_{t+1}), P(d_t)) = \varepsilon(\tilde \nabla f(x_{t+1}), \tilde \nabla f(x_{t+1}^{theo})) + \varepsilon(\tilde \nabla f(x_{t+1}^{theo}), d_t) + \varepsilon(d_t, P(d_t)).
    \]
    \begin{enumerate}
      \item Without strong assumptions on \(f\), the first term
    (being a difference between two subgradients)
       is of the order of  \(O(1/\sqrt t)\).
      \item The second term can be controlled by \(O(1/\sqrt t)\) when \(d_t = \tilde \nabla f(x_t)\) (subgradient boosting) and \(0\) if \(d_t\) is the proximal direction (proximal boosting).
      \item The third term usually requires the edge hypothesis (as used in the literature and in this paper) and a specific mechanism inside the algorithm to be controlled,
      such as the residual inspired by \citep{grubb_generalized_2011}.
    \end{enumerate}
}

\neurocomp{
    Overall,  even if proximal boosting benefits from the cancellation of the second error term,  the first and the third ones remain limiting, resulting in an \(O(1/\sqrt t)\) rate,  as in the case of subgradient boosting.
}

\end{remark}

	\section{Numerical analysis}
		\label{sec:numerical}
In Section~\ref{sec:algorithm}, proximal boosting algorithms have been introduced in a fairly general way.
However, the empirical results presented in this section are based on an implementation (See Algorithm~\ref{alg:prox_meta}) incorporating some modifications that have made the success of gradient boosting.


\begin{algorithm}[ht]
  \begin{algorithmic}[1]
    \REQUIRE
    $\nu \in (0, 1]$ (shrinkage coefficient), $\lambda > 0$ (proximal step).
    \STATE Set $f_0 \in \argmin_{g \in \F_0} C(g)$, \(\Delta_0 \gets 0\) \COMMENT{initialization}.
    \FOR{$t=0$ \TO $T-1$}
      \STATE \neurocomp{Pseudo-residuals \underline{(see Appendix~\ref{app:losses})}:
      \[
        \begin{cases}
          r \gets -\fsub C (f_t) & \text{ for gradient boosting}, \\
          r \gets -\proxdir{\lambda} C(f_t) & \text{ for proximal boosting}.
        \end{cases}
      \]}
      \STATE \neurocomp{Regression of the vector \(r + \Delta_t \in \R^n\) onto \((X_1, \dots, X_n)\):}
      $$g_{t+1} \in \argmin_{g \in \F} \norm{g(X_1^n) - (r + \Delta_t)}.$$
      \STATE \neurocomp{Residual:}
      \[
        \begin{cases}
          \Delta_{t+1} \gets 0 & \text{ for vanilla boosting}, \\
          \Delta_{t+1} \gets r + \Delta_t - g_{t+1}(X_1^n) & \text{ for residual boosting}.
        \end{cases}
      \]
      \STATE \neurocomp{Line-search \underline{(see Appendix~\ref{app:losses})}:}
      $$\gamma_{t+1} \in \argmin_{\gamma \in \R} C(f_t + \gamma g_{t+1}).$$
      \STATE \neurocomp{Update:}
      $$f_{t+1} \gets f_t + \nu \gamma_{t+1} g_{t+1}.$$
    \ENDFOR
    \ENSURE $f_T$.
  \end{algorithmic}
  \caption{Meta-algorithm for boosting.}
  \label{alg:prox_meta}
\end{algorithm}

First of all, the proximal step is fixed to some positive value: \(\lambda_t = \lambda > 0\);
and the update rule \(f_{t+1} \gets f_t + \lambda_{t+1} g_{t+1}\) is replaced by \(f_{t+1} \gets f_t + \nu \gamma_{t+1} g_{t+1}\), where
\[
  \gamma_{t+1} \in \argmin_{\gamma \in \R} C(f_t + \gamma g_{t+1}).
\]
In other words, the step size is tuned by a shrinkage coefficient (or learning rate) \(\nu \in (0, 1]\) and a line search producing the largest decrease of the objective function.

The learning rate is known to be a key element of boosting machines in order to obtain a good generalization performance.
To understand that fact, let us remark that the number of iterations $T$ acts on two regularization mechanisms.
The first one is statistical ($T$ controls the complexity of the subspace in which $f_T$ lies) and the second one is numerical ($T$ controls the precision to which the empirical risk $C$ is minimized).
The shrinkage coefficient $\nu$ tunes the balance between these two regularization mechanisms.

Besides the learning rate, the step size is controlled by a line search, that simply scales the weak learner $g_{t+1}$ by a constant factor.
Actually, since the class of weak learners \(\F\) is in practice a set of regression trees (implemented in Scikit-learn \citep{pedregosa_scikit-learn:_2011}), a multiple line search is used, as proposed by \citet{friedman_greedy_2001}: a line search is performed sequentially for each leaf of the decision tree, such that each level of the piecewise constant function $g_{t+1}$ is scaled with its own factor.
All variants of proximal and gradient boosting are implemented based on Algorithm~\ref{alg:prox_meta} in the Scikit-learn fashion \citep{pedregosa_scikit-learn:_2011} and are freely available in the Python package \texttt{optboosting}\footnote{\url{https://github.com/msangnier/optboosting}}.

\subsection{Behavior of proximal boosting}
  Based on synthetic data, this section aims at numerically illustrating the performance of proximal boosting compared to gradient boosting.
  For this purpose, two synthetic models are studied, both coming from \citet{biau_accelerated_2018,biau_cobra:_2016}:
  \begin{description}
    \item[Regression:]
    \[
      \left\|
      \begin{array}{l}
        n=800, d=100;\\
        Y=-\sin(2X^{(1)})+{X^{(2)}}^2+X^{(3)}-\exp(-X^{(4)})+Z_{0.5},
      \end{array}
      \right.
    \]
    \item[Classification:]
    \[
      \left\|
      \begin{array}{l}
        n=1500, d=50;\\
        Y=
        \begin{cases}
          1 & \text{if } X^{(1)}+{X^{(4)}}^3+X^{(9)}+\sin(X^{(12)}X^{(18)})+Z_{0.1}>0.38; \\
          -1 & \text{otherwise},
        \end{cases}
      \end{array}
      \right.
    \]
  \end{description}
   \neurocomp{where $Z_{\sigma^2}$ is a random variable independent from \(X\), following a normal distribution with zero mean and variance \(\sigma^2\).}

  The first model covers an additive regression problem, while the second covers a binary classification task with covariate interactions.
  In both cases, we consider an input random variable $X \in \R^d$, the covariate of which, denoted $(X^{(j)})_{1 \le j \le d}$, are normally distributed with zero mean and covariance matrix $\Sigma = \left( 2^{-|i-j|} \right)_{1 \le i, j \le d}$.
  Moreover, in these synthetic models of regression and classification, an additive and independent noise
  is embodied by the random variable $Z_{\sigma^2}$.

  Four different losses are considered (see Table \ref{tab:losses} for a brief description):
  least squares and least absolute deviations for regression; exponential (with $\beta=1$) and hinge for classification.
  Computations for the corresponding (sub)gradients and proximal operators are detailed in Appendix~\ref{app:losses}.
  On that occasion, it can be remarked that the direction of descent $\proxdir{\lambda} C(f_t)$ of proximal boosting applied with the least squares loss is the same as that of gradient boosting, $\fgrad C(f_t)$, up to a constant factor (see Appendix~\ref{app:losses}).
  In other words, proximal and gradient boosting are exactly equivalent.

  In addition, note that we also considered other kind of losses such as the pinball loss for regression and the logistic loss for classification (see Table \ref{tab:losses}).
  Nevertheless, since the numerical behaviors are respectively very close to the least absolute deviations and the exponential cases, the results are not reported.

  \begin{table}
    \center
    \begin{tabular}{p{2.3cm}lll}
      \textsc{Loss} & \textsc{Parameter} & $\ell(y, y')$ & \textsc{Type}\\ \hline
      Least squares & - & $(y-y')^2/2$  & Regression \\
      Least absolute deviations & - & $|y-y'|$ & Regression \\
      Pinball & $\tau \in (0, 1)$ & $\max(\tau(y-y'), (\tau-1)(y-y'))$ & Regression \\
      Exponential & $\beta > 0$ & $\exp(-\beta yy')$ & Classification \\
      Logistic & - & $\log_2(1 + \exp(-yy'))$ & Classification \\
      Hinge & - & $\max(0, 1-yy')$ & Classification
    \end{tabular}
    \caption{Loss functions.}
    \label{tab:losses}
  \end{table}

  In the following numerical experiments, the random sample generated based on each model is divided into a training set (50\%) to fit the method and a test set (50\%).
  The performance of the methods are appraised through several curves representing the training and test losses along the $T=1000$ iterations of boosting.

  \subsubsection{Convergence}
\begin{figure}
      \center
      \includegraphics[height=0.35\textwidth]{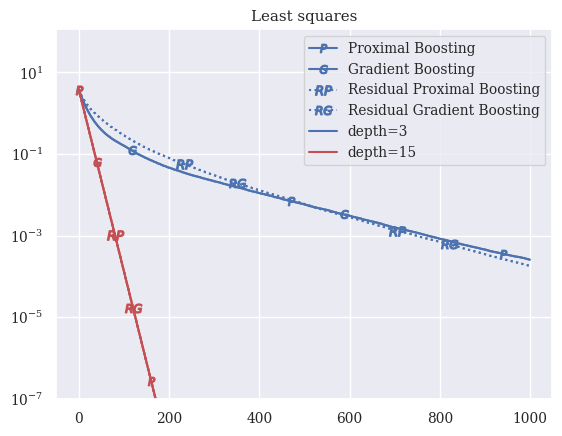}
      \includegraphics[height=0.35\textwidth]{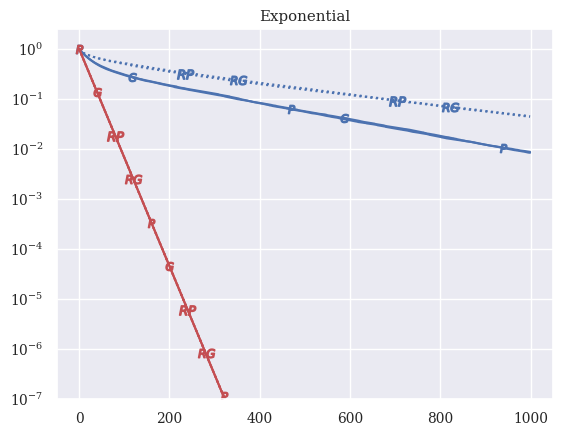}
      \includegraphics[height=0.35\textwidth]{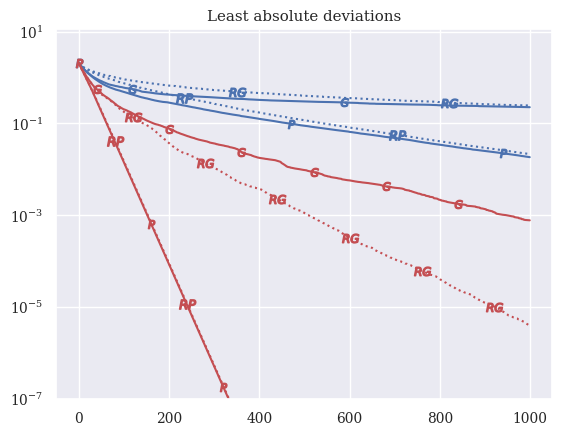}
      \includegraphics[height=0.35\textwidth]{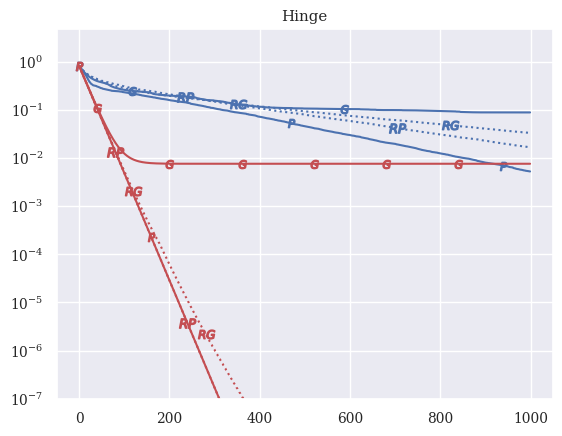}
      \caption{\label{fig:depth}
      Training losses for two values of maximal depth (3 in blue, 15 in red) \vs number of iterations on the horizontal axis.}
    \end{figure}

    As a first numerical experiment, we aim at illustrating the convergence of proximal boosting (see Section~\ref{sec:cv}) for two classes $\mathcal F$ of weak learners: regression trees with maximal depth $3$ (in blue in Figure~\ref{fig:depth}) and with maximal depth $15$ (in red in Figure~\ref{fig:depth}).
    This last class of weak learners is supposed to make almost no error in approximating the directions of descent, thus leading to quasi-standard optimization algorithms.

    For the purpose of the analysis, parameters $\lambda$ and $\nu$ are set to standard values: $\lambda = 1$, $\nu = 5 \cdot 10^{-2}$, which does not hurt the generality of the forthcoming interpretations.
    Moreover, gradient boosting and its variant proposed by \citet{grubb_generalized_2011}, residual gradient boosting, are included as references.

    Let us analyze the top panels of Figure~\ref{fig:depth}: for differentiable losses (least squares and exponential), proximal and gradient descents behave exactly the same
    (\neurocomp{curves with symbols \textbf{P} and \textbf{G} are mixed up}).
    Moreover, as theoretically analyzed in Theorem~\ref{thm:cv_results}, the rate of convergence of proximal boosting is linear with a slope that increases with the capacity of the class of weak learners (even though the exponential loss is not strongly convex).

    Still for differentiable losses, the use of the residual originally introduced to derive a convergence rate under weak assumptions (represented with dotted lines \neurocomp{and symbols \textbf{RP} and \textbf{RG}} in Figure~\ref{fig:depth}) does not seem to help convergence neither with a large class of weak learners (in red, the residual is in fact always almost null), nor with a restricted class (in blue).

    Concerning non-differentiable losses (least absolute deviations and hinge on the bottom panels of Figure~\ref{fig:depth}), proximal boosting converges faster than gradient boosting, which does not seem to converge for the hinge loss.
    In addition, it is noticeable to observe that convergence of proximal boosting seems almost linear while the empirical risk violates the assumptions of smoothness required for Theorem~\ref{thm:cv_results}.

    For non-differentiable losses, the use of the residual helps gradient boosting to converge.
    Yet, we remark that residual proximal boosting behaves similarly to proximal boosting (curves \neurocomp{with symbols \textbf{RP} and \textbf{P}} are mixed up),
    \new{suggesting that, from a convergence point of view, this mechanism is more needed for a theoretical purpose than for a practical one.}

    \begin{figure}
      \center
      \includegraphics[height=0.35\textwidth]{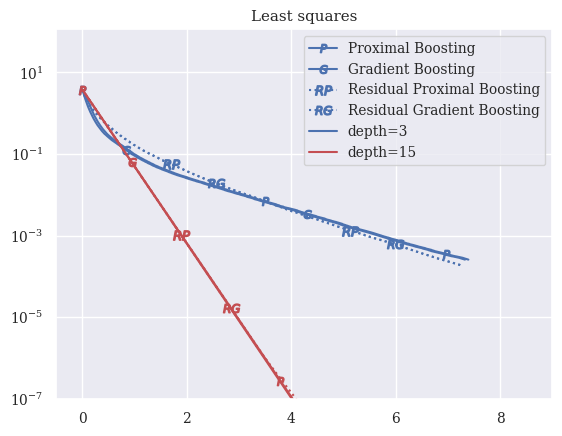}
      \includegraphics[height=0.35\textwidth]{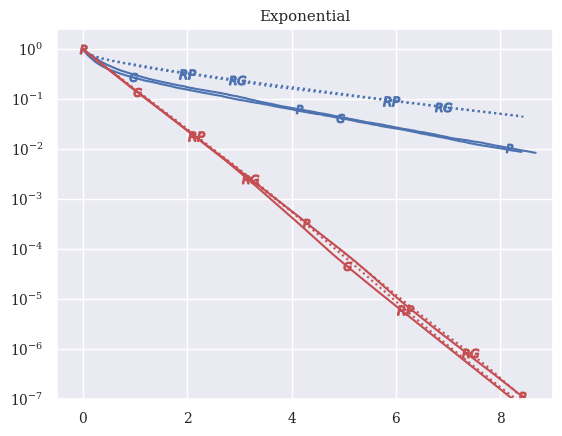}
      \includegraphics[height=0.35\textwidth]{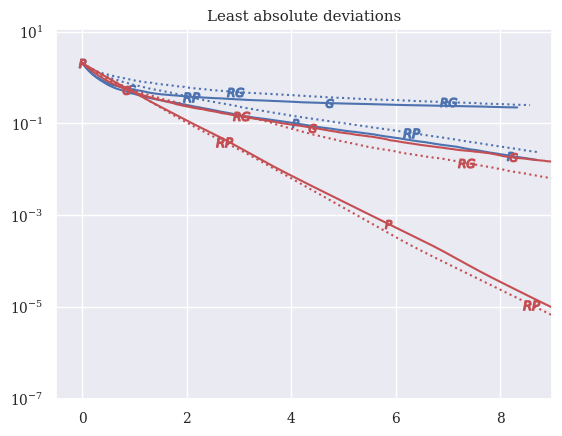}
      \includegraphics[height=0.35\textwidth]{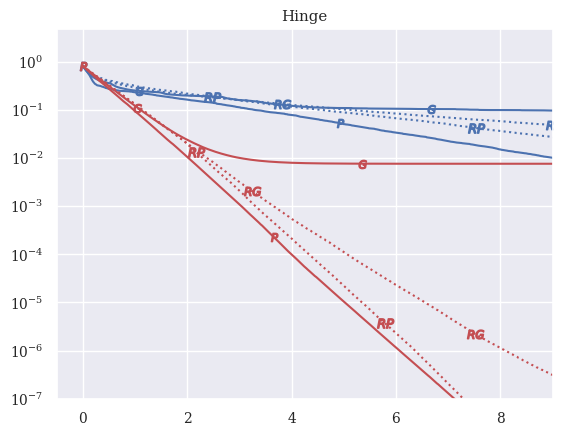}
      \caption{\label{fig:depth_clock}
      \neurocomp{Training losses for two values of maximal depth (3 in blue, 15 in red) \vs clock time (seconds) on the horizontal axis.}}
    \end{figure}

    \neurocomp{As a last piece of evidence, Figure~\ref{fig:depth_clock} depicts the same experiment as the previous one but with the clock time on the horizontal axis.
    We can remark that the behavior of algorithms is similar when convergence is analyzed with respect to the number of iterations or to the time elapsed.
    This shows that computing a proximal direction of descent is not more expensive than computing a (sub)gradient, which is in favor of proximal boosting.}

    \new{Overall, this numerical experiment confirms the initial intuition that proximal boosting behaves better than gradient boosting and residual gradient boosting in the non-differentiable cases.}
    Keeping in mind that behaviors are similar for differentiable losses, we carry on the study only with least absolute deviations and hinge losses.

  \subsubsection{Proximal step}
\begin{figure}
      \center
      \includegraphics[height=0.35\textwidth]{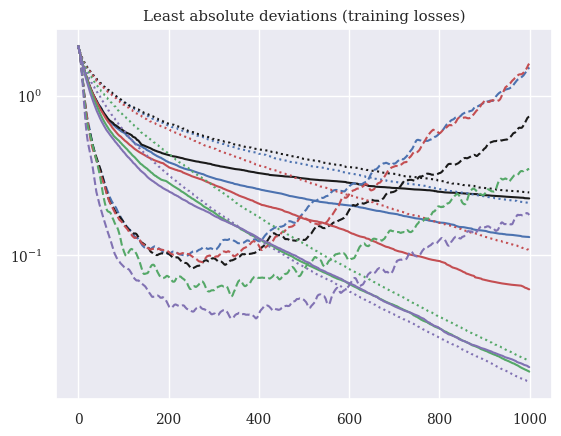}
      \includegraphics[height=0.35\textwidth]{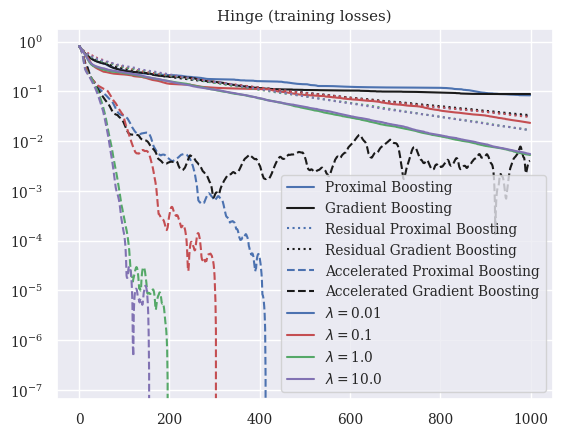}
      \includegraphics[height=0.35\textwidth]{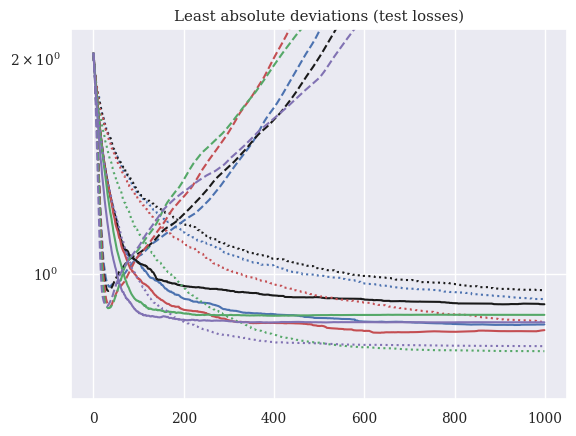}
      \includegraphics[height=0.35\textwidth]{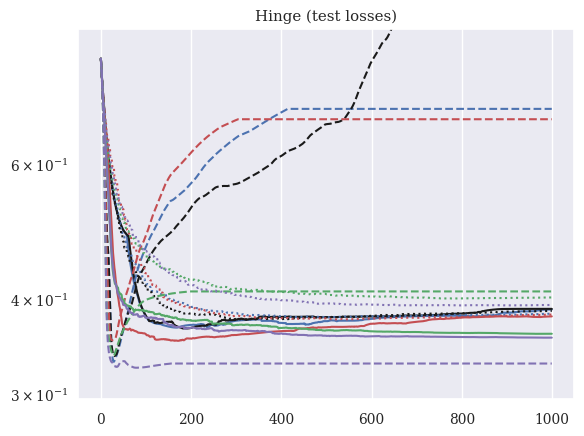}
      \caption{\label{fig:step}
      Training (top) and test (bottom) losses of proximal boosting algorithms for several values of the proximal step \(\lambda\) \vs number of iterations on the horizontal axis.}
    \end{figure}

    We aim at illustrating the impact of the proximal step \(\lambda\) intervening in proximal boosting as a new parameter.
    For this purpose, Figure~\ref{fig:step} depicts the trend of training (top) and test (bottom) losses of proximal boosting for $\lambda \in \{10^{-2},10^{-1}, \dots, 10^2\}$ (see the different colors) and decision trees of maximal depth \(3\) as weak learners.
    Compared algorithms include proximal (Algorithm~\ref{alg:prox}), residual proximal (Algorithm~\ref{alg:prox_res}) and accelerated proximal (Algorithm~\ref{alg:prox_acc}) boosting,
    as well as their gradient counterparts (in black, independent of \(\lambda\)).

    Figure~\ref{fig:step} shines a light of the tie between the proximal step and the convergence rate: the bigger $\lambda$, the faster the convergence of the training and test losses.
    As a consequence (see the top panel), proximal boosting prevails over gradient boosting from an optimization perspective because it converges faster for sufficiently large \(\lambda\).
    Regarding the training loss, the advantage of using the residual is not clear since proximal boosting offers similar convergence rates than residual proximal boosting for large values of \(\lambda\), and converges faster than residual gradient boosting.

    Analyzing the test loss, proximal and residual proximal boosting achieve lower errors than gradient and residual gradient boosting for intermediate values of \(\lambda\) (between \(0.1\) and \(10\)).
    In addition, their behavior is quite stable with respect to the parameters from our experience.

    From all points of view, using a proximal direction of descent is a real advantage over subgradient.
    Besides, from a global perspective, proximal boosting helps to build more accurate models than gradient boosting.

    \new{
    This numerical experiment is also a place for studying the benefit of incorporating Nesterov's acceleration into boosting.
    For proximal as well as gradient boosting, acceleration speeds up the decrease of the training and test losses, and thus it makes it possible to build boosted models with very few weak learners.
    Nevertheless, both accelerated boosting approaches suffer from instabilities leading to divergence, on the training and on the test sets.
    Regarding the test error, they do not seem to be capable to produce very accurate models (on the bottom panel of Figure~\ref{fig:step}, non-accelerated methods offer a lower test error than accelerated ones).
    Remark that, accelerated proximal boosting performs definitely better than its accelerated gradient counterpart.
    We guess that
    \begin{enumerate}
      \item the procedure is diverging because Nesterov's extrapolation intensifies the boosting approximation error made at each iteration;
      \item the acceleration makes the method very sensitive and dependent on a fine tuning procedure in order to perform well in generalization.
    \end{enumerate}
    Even though divergence on the training error seems to always occur in the overfitting regime (\ie after the minimal test error, see Figure~\ref{fig:step}), these observations are not in line with a statistically reliable learning technique.
    As a consequence, such methods are only recommended to build models with very few trees, for instance because of hardware constraints.}


\subsection{Generalization in real world cases}
\label{sec:real_world}
  This section aims at comparing the generalization ability of the proposed boosting estimators with respect to variants of gradient boosting, as well as extreme gradient boosting (XGBoost) \citep{chen_xgboost:_2016} and random forests \citep{breiman_random_2001}.
  The last two methods are introduced in the numerical comparison only as benchmarks.
  Indeed, random forests aggregate weak learners but with equal weights, and XGBoost is a boosting method based on second order optimization.
  From a strict optimization point of view, second order optimization is not applicable to non-differentiable loss functions, nevertheless, given the liberty taken with Nesterov's acceleration, XGBoost is applied as a black box for minimizing the empirical loss.
  It is important to point out that, up to our knowledge, there is no convergence result for XGBoost with non-differentiable losses.

  Comparison is based on nine datasets (available on the UCI Machine Learning repository), the characteristics of which are described in Table~\ref{tab:datasets}.
  The first six are univariate regression datasets, while the three others relate to binary classification problems.
  In both situations, the sample is split into a training set (50\%), a validation set (25\%) and a test set (25\%).
  The parameters of the methods (number of weak classifiers $T \in [1, 1000]$, maximal depth of decision trees varying in $[1, 3, 5]$, learning rate $\nu \in \{5 \cdot 10^{-2}, 10^{-1}, 3 \cdot 10^{-1}, 5 \cdot 10^{-1}, 1\}$ and proximal step $\lambda \in \{10^{-3}, 10^{-2}, \dots, 10^2\}$ for boosting, completed with
  the maximal number of features for random forests) are selected as minimizers of the loss computed on the validation set for models fitted on the training set.
  Then, models are refitted on the training and the validation sets with selected parameters.
  Finally, the generalization ability of the methods is estimated by computing the loss (and the misclassification rate for classification models) on the test set.
  These quantities are reported through statistics computed on 20 random splits of the datasets.

  \begin{table}[H]
    \center
    \begin{tabular}{lccl}
      \textsc{Dataset} & $n$ & $d$ & \textsc{Type}\\ \hline
      Whitewine & \(4898\) & \(11\) & Regression \\
      Redwine & \(1599\) & \(11\) & Regression \\
      BostonHousing & \(506\) & \(13\) & Regression \\
      Crabs & \(200\) & \(4\) & Regression \\
      Engel & \(235\) & \(1\) & Regression \\
      Sniffer & \(125\) & \(4\) & Regression \\
      Adult & $30162$ & $13$ & Classification \\
      Advertisements & $2359$ & $1558$ & Classification \\
      Spam & $4601$ & $57$ & Classification \\
    \end{tabular}
    \caption{Real-world datasets (\(n\): sample size, \(d\): number of attributes).}
    \label{tab:datasets}
  \end{table}

  The losses considered in these experiments are least squares, least absolute deviations and pinball (with $\tau=0.9$) for the regression problems,
  as well as exponential (with $\beta=1$) and hinge for the classification tasks (see Table~\ref{tab:losses} for a quick definition and Appendix~\ref{app:losses} for the details).
  Since random forests are not expli\-citly designed for minimizing theses losses, only the least squares test loss and the classification error are reported.

  \subsubsection{Regression problems}
\begin{figure}
      \center
      \includegraphics[width=.9\textwidth]{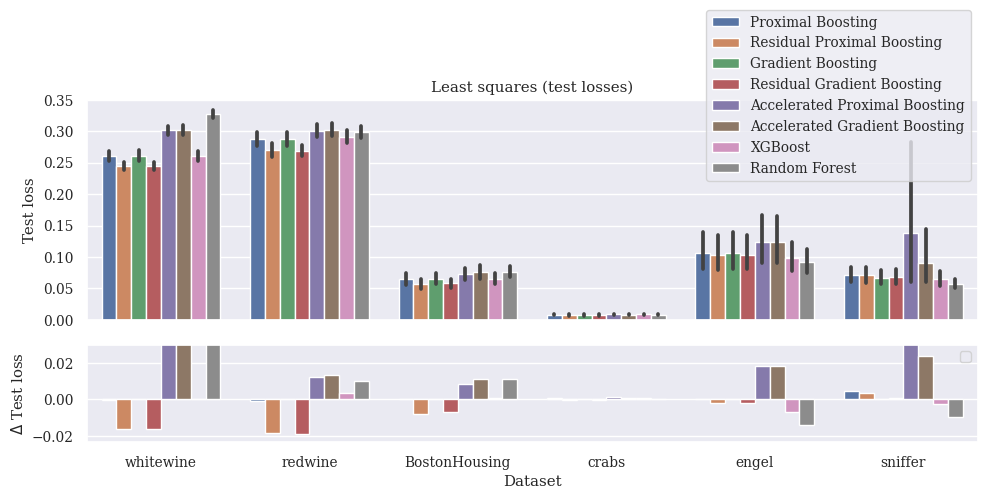}
      \includegraphics[width=.9\textwidth]{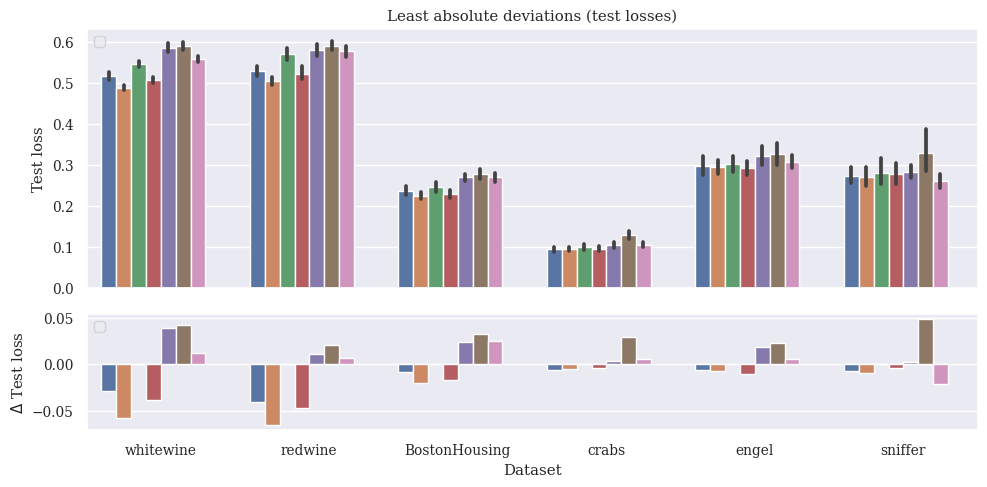}
      \includegraphics[width=.9\textwidth]{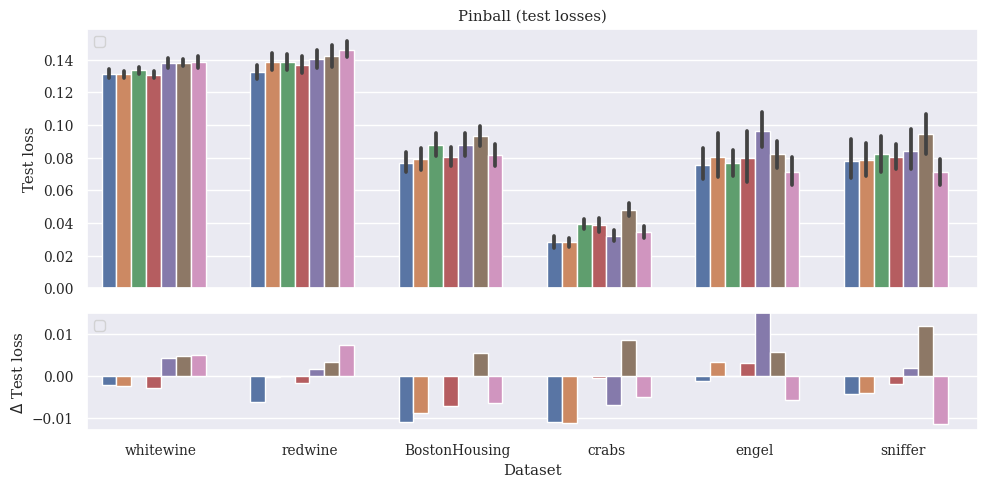}
      \caption{Losses on test datasets for the least squares (top), least absolute deviations (middle) and pinball (bottom) losses.
      \emph{\(\Delta\) Test loss} refers to the increment of the loss from that of gradient boosting.
      The methods proposed in this article are in blue, orange and green.}
      \label{fig:regression_loss}
    \end{figure}

    Test losses for the least squares (top), least absolute deviations (middle) and pinball (bottom) losses are described in Figure~\ref{fig:regression_loss}.
    \emph{\(\Delta\) Test loss} refers to the increment of the loss from that of gradient boosting.

    Regarding the least squares setting, let us remind that gradient and proximal boosting boil down to be the same method (the directions of descent are exactly the same).
    We observe that they achieve a performance comparable to extreme gradient boosting and better than random forests.
    Moreover, even though residual boosting was not designed for differentiable losses, it provides the most accurate models for 3 datasets out of 6.

    Looking now at least absolute deviations and pinball losses,
    we observe that proximal boosting always achieves better predictions than gradient boosting.
    In addition, in the bulk of the situations, the most accurate method is either proximal or residual proximal boosting.
    This confirms our intuition concerning the need for optimization techniques suited for non-differentiable loss functions.

    \new{Regarding accelerated versions of boosting, as expected they do not produce more accurate models than vanilla boosting, very likely because convergence is so fast that tuning parameters becomes excessively tricky.
    Incidentally, we remark that accelerated proximal boosting offers better generalization performances than accelerated gradient boosting for non-differentiable losses (except for the dataset Engel).}

  \subsubsection{Classification problems}
\begin{figure}
      \center
      \includegraphics[width=\textwidth]{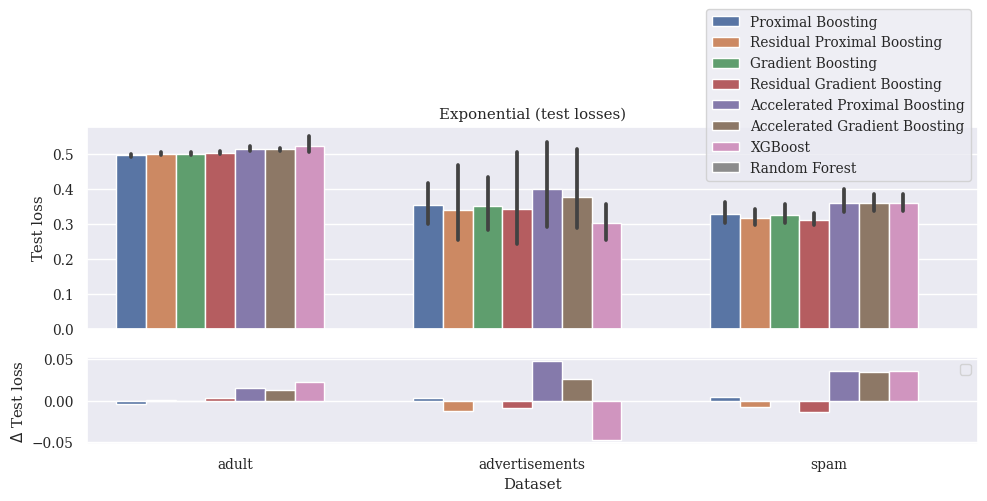}
      \includegraphics[width=\textwidth]{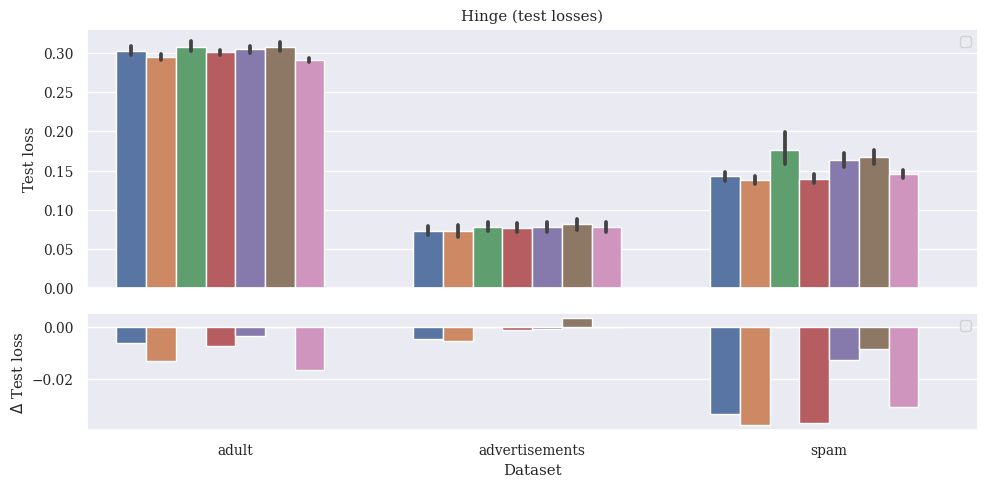}
      \caption{Losses on test datasets for the exponential (top) and hinge (bottom) losses.
      \emph{\(\Delta\) Test loss} refers to the increment of the loss from that of gradient boosting.
      The methods proposed in this article are in blue, orange and green.}
      \label{fig:classification_loss}
    \end{figure}

\begin{figure}
      \center
      \includegraphics[width=\textwidth]{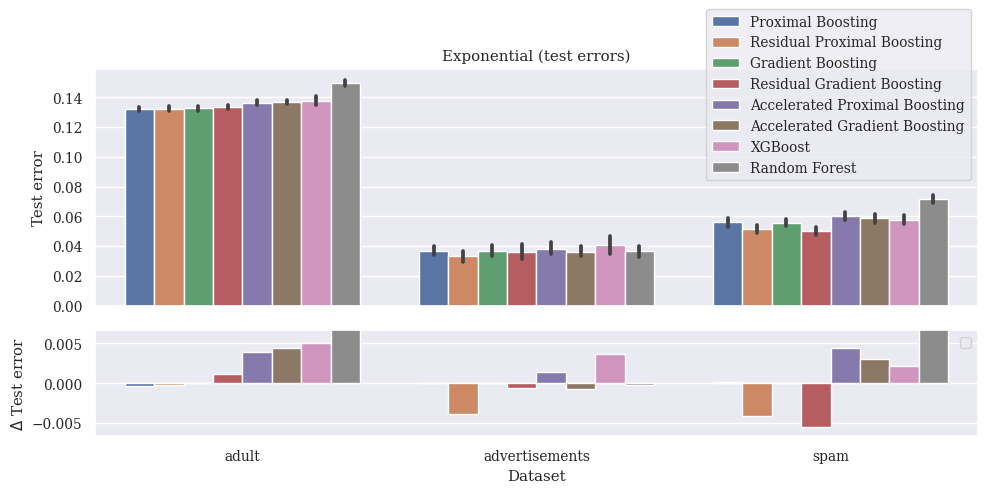}
      \includegraphics[width=\textwidth]{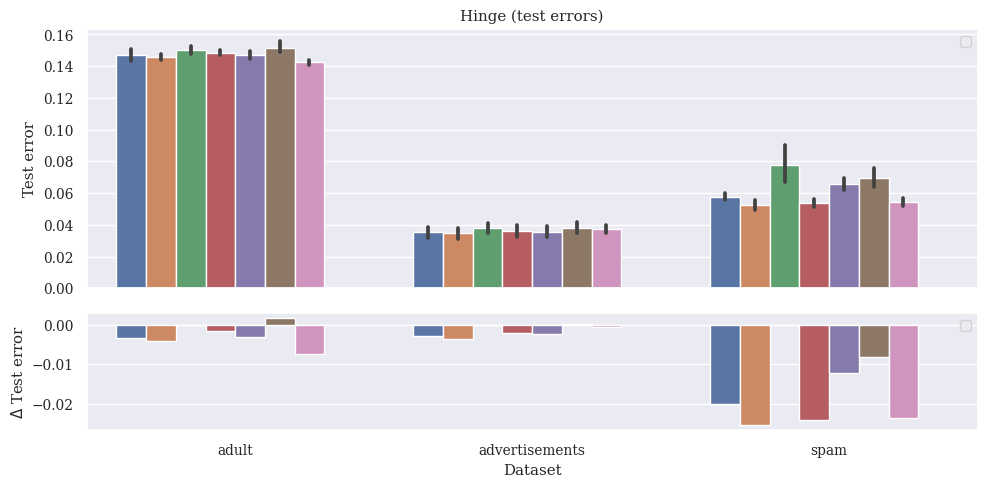}
      \includegraphics[width=\textwidth]{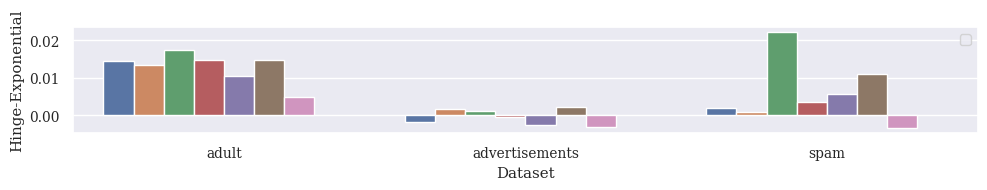}
      \caption{Misclassification rates on test datasets for the exponential (top) and hinge (bottom) losses.
      \emph{\(\Delta\) Test error} and \emph{Hinge-Exponential} refer to the increment of the misclassification rate respectively from that of gradient boosting and from that obtained with the exponential loss.
      The methods proposed in this article are in blue, orange and green.}
      \label{fig:classification_error}
    \end{figure}

    Losses and misclassification rates computed on the test datasets are depicted respectively in Figure~\ref{fig:classification_loss} and in Figure~\ref{fig:classification_error} for the exponential (top) and the hinge (bottom) losses.
    Besides \emph{\(\Delta\) Test loss/error}, referring to the increment of the loss or misclassification rate from that of gradient boosting, \emph{Hinge-Exponential} in Figure~\ref{fig:classification_error} represents the increment of the misclassification rate of hinge loss-based boosting from that obtained with the exponential loss.

    Regarding both indicators (loss in Figure~\ref{fig:classification_loss} and error in Figure~\ref{fig:classification_error}), four methods share the winners' podium: proximal boosting (blue), residual proximal boosting (orange), residual gradient boosting (red) and XGBoost (pink).
    For the hinge loss, proximal or residual proximal boosting are always better than gradient and residual gradient boosting.
    Moreover, accelerated proximal boosting (purple) always gives better loss and accuracy than gradient and accelerated gradient boosting (brown).
    Both observations confirm the interest of proximal-based boosting for non-differentiable losses.

    It is remarkable that XGBoost performs quite well with the hinge loss, while it was not originally designed for non-differentiable losses.
    Nevertheless, the bottom panel of Figure~\ref{fig:classification_error} shows that, overall using a hinge loss instead of an exponential loss is rarely a big advantage, except to obtain sporadically a marginal gain in accuracy.


	\section{Conclusion}

Building upon the proximal point method for convex and non-smooth optimization, this paper has introduced two novel boosting algorithms, nicknamed \emph{proximal boosting} and \emph{residual proximal boosting}, which have appeal for non-differentiable loss functions $\ell$.
A theoretical study demonstrates convergence of proximal and residual proximal boosting from an optimization point of view (under different hypotheses on the loss function).
Numerical experiments on synthetic data confirm the theoretical convergence results and show a significant impact of the newly introduced parameter $\lambda$.
Correctly tuned, this parameter provides a noticeable improvement of proximal-based boosting over gradient-based boosting for non-differentiable loss function, from both the optimization and the statistical points of view.
Moreover, in real-world regression and classification situations, proximal or residual proximal boosting often achieve the best test loss and are, overall, very competitive with state-of-the-art boosting approaches.

\new{As a by-product, we have also studied incorporating Nesterov's acceleration to proximal boosting, as done with gradient boosting in \citep{biau_accelerated_2018}.
Numerically, we observe instabilities in both algorithms, leading to divergence on the training and the test sets.
Our experience is that accelerated boosting is very sensitive to hyperparameters and thus tricky to tune.
Despite the fact that these procedures rarely provide good generalization results, accelerated proximal boosting seems to perform better than its gradient counterpart for non-differentiable losses.}

\neurocomp{
Going further in the theoretical analysis of accelerated proximal boosting is also an exciting perspective.  In particular,  \citet{lu_accelerating_2020} recently proposed a variant of accelerated gradient boosting \citep{biau_accelerated_2018} with guaranteed convergence for differentiable and smooth losses.
Establishing similar results for accelerated proximal boosting constitutes an important challenge both from a numerical and a theoretical point of view. 
}

On another note,  we believe that the connection between boosting and functional optimization can be much more investigated.
In particular, advances in optimization theory can spread to boosting, just like the Frank-Wolfe algorithm has impacted boosting \citep{wang_functional_2015,jaggi_revisiting_2013}.
This may also hold true for non-differentiable and non-convex optimization (see for instance \citep{ochs_ipiano:_2014}).

	\section*{Acknowledgements}
		The authors are thankful to Gérard Biau and Jalal Fadili for enlightening discussions.
		They also thank the three anonymous referees for their constructive comments.

	\bibliography{maxime,claire,supp}

\begin{thebibliography}{54}
\providecommand{\natexlab}[1]{#1}
\providecommand{\url}[1]{\texttt{#1}}
\expandafter\ifx\csname urlstyle\endcsname\relax
  \providecommand{\doi}[1]{doi: #1}\else
  \providecommand{\doi}{doi: \begingroup \urlstyle{rm}\Url}\fi

\bibitem[Ahamad et~al.(2020)Ahamad, Aktar, Rashed-Al-Mahfuz, Uddin, Liò, Xu,
  Summers, Quinn, and Moni]{ahamad_machine_2020}
Md.~M. Ahamad, S.~Aktar, Md. Rashed-Al-Mahfuz, S.~Uddin, P.~Liò, H.~Xu, M.~A.
  Summers, Julian M.~W. Quinn, and M.~A. Moni.
\newblock A machine learning model to identify early stage symptoms of
  {SARS}-{Cov}-2 infected patients.
\newblock \emph{Expert Systems with Applications}, 160:\penalty0 113661, 2020.

\bibitem[Awal et~al.(2021)Awal, Masud, Hossain, Bulbul, Mahmud, and
  Bairagi]{awal_novel_2021}
Md.~A. Awal, M.~Masud, Md.~S. Hossain, A.~A.-M. Bulbul, S.~M.~H. Mahmud, and
  A.~K. Bairagi.
\newblock A {Novel} {Bayesian} {Optimization}-{Based} {Machine} {Learning}
  {Framework} for {COVID}-19 {Detection} {From} {Inpatient} {Facility} {Data}.
\newblock \emph{IEEE Access}, 9:\penalty0 10263--10281, 2021.

\bibitem[Beck and Teboulle(2009)]{beck2009fast}
A.~Beck and M.~Teboulle.
\newblock A fast iterative shrinkage-thresholding algorithm for linear inverse
  problems.
\newblock \emph{SIAM {J}ournal on {I}maging {S}ciences}, 2\penalty0
  (1):\penalty0 183--202, 2009.

\bibitem[Biau and Cadre(2021)]{biau_optimization_2021}
G.~Biau and B.~Cadre.
\newblock \emph{Optimization by Gradient Boosting}, pages 23--44.
\newblock Springer International Publishing, Cham, 2021.

\bibitem[Biau et~al.(2016)Biau, Fischer, Guedj, and Malley]{biau_cobra:_2016}
G.~Biau, A.~Fischer, B.~Guedj, and J.D. Malley.
\newblock {COBRA}: {A} combined regression strategy.
\newblock \emph{Journal of Multivariate Analysis}, 146:\penalty0 18--28, 2016.

\bibitem[Biau et~al.(2019)Biau, Cadre, and Rouvière]{biau_accelerated_2018}
G.~Biau, B.~Cadre, and L.~Rouvière.
\newblock {Accelerated} {Gradient} {Boosting}.
\newblock \emph{Machine Learning}, 108\penalty0 (6):\penalty0 971--992, 2019.
\newblock ISSN 0885-6125.

\bibitem[Breiman(1997)]{breiman_arcing_1997}
L.~Breiman.
\newblock Arcing the {Edge}.
\newblock Technical {Report} 486, Statistics Department, University of
  California, Berkeley, 1997.

\bibitem[Breiman(1998)]{breiman_arcing_1998}
L.~Breiman.
\newblock Arcing classifier (with discussion and a rejoinder by the author).
\newblock \emph{The Annals of Statistics}, 26\penalty0 (3):\penalty0 801--849,
  1998.

\bibitem[Breiman(1999)]{breiman_prediction_1999}
L.~Breiman.
\newblock Prediction {Games} and {Arcing} {Algorithms}.
\newblock \emph{Neural Computation}, 11\penalty0 (7):\penalty0 1493--1517,
  1999.

\bibitem[Breiman(2000)]{breiman_infinite_2000}
L.~Breiman.
\newblock Some {Infinite} {Theory} for {Predictor} {Ensembles}.
\newblock Technical {Report} 577, Statistics Department, University of
  California, Berkeley, 2000.

\bibitem[Breiman(2001)]{breiman_random_2001}
L.~Breiman.
\newblock Random {Forests}.
\newblock \emph{Machine Learning}, 45\penalty0 (1):\penalty0 5--32, 2001.

\bibitem[Breiman(2004)]{breiman_population_2004}
L.~Breiman.
\newblock Population theory for boosting ensembles.
\newblock \emph{The Annals of Statistics}, 32\penalty0 (1):\penalty0 1--11,
  2004.

\bibitem[Bühlmann and Hothorn(2007)]{buhlmann_boosting_2007}
P.~Bühlmann and T.~Hothorn.
\newblock Boosting {Algorithms}: {Regularization}, {Prediction} and {Model}
  {Fitting}.
\newblock \emph{Statistical Science}, 22\penalty0 (4):\penalty0 477--505, 2007.

\bibitem[Bühlmann and Yu(2003)]{buhlmann_boosting_2003}
P.~Bühlmann and B.~Yu.
\newblock Boosting {With} the {L2} {Loss}.
\newblock \emph{Journal of the American Statistical Association}, 98\penalty0
  (462):\penalty0 324--339, 2003.

\bibitem[Cai et~al.(2020)Cai, Hang, Yang, and Lin]{cai_boosted_2020}
Y.~Cai, H.~Hang, H.~Yang, and Z.~Lin.
\newblock Boosted {Histogram} {Transform} for {Regression}.
\newblock In \emph{Proceedings of the 37th {International} {Conference} on
  {Machine} {Learning}}, pages 1251--1261. PMLR, 2020.

\bibitem[Chen et~al.(2022)Chen, Chu, Zhao, Luo, and Luo]{chen_output_2022}
J.~Chen, Z.~Chu, R.~Zhao, A.~F. Luo, and K.~H. Luo.
\newblock Output prediction of alpha-type {Stirling} engines using gradient
  boosted regression trees and corresponding heat recovery system optimization
  based on improved {NSGA}-{II}.
\newblock \emph{Energy Reports}, 8:\penalty0 835--846, 2022.

\bibitem[Chen and Guestrin(2016)]{chen_xgboost:_2016}
T.~Chen and C.~Guestrin.
\newblock {XGBoost}: {A} {Scalable} {Tree} {Boosting} {System}.
\newblock In \emph{{Proceedings} of the 22nd {ACM} {SIGKDD} {International}
  {Conference} on {Knowledge} {Discovery} and {Data} {Mining}}, pages 785--794,
  New York, NY, USA, 2016. ACM.

\bibitem[Combettes and Wajs(2005)]{combettes_signal_2005}
P.~Combettes and V.~Wajs.
\newblock Signal {Recovery} by {Proximal} {Forward}-{Backward} {Splitting}.
\newblock \emph{Multiscale Modeling \& Simulation}, 4\penalty0 (4):\penalty0
  1168--1200, 2005.

\bibitem[Cui et~al.(2021)Cui, Hang, Wang, and Lin]{cui_gbht_2021}
J.~Cui, H.~Hang, Y.~Wang, and Z.~Lin.
\newblock {GBHT}: {Gradient} {Boosting} {Histogram} {Transform} for {Density}
  {Estimation}.
\newblock In \emph{Proceedings of the 38th {International} {Conference} on
  {Machine} {Learning}}, pages 2233--2243. PMLR, 2021.

\bibitem[Freund(1995)]{freund_boosting_1995}
Y.~Freund.
\newblock Boosting a {Weak} {Learning} {Algorithm} by {Majority}.
\newblock \emph{Information and Computation}, 121\penalty0 (2):\penalty0
  256--285, 1995.

\bibitem[Freund and Schapire(1996)]{freund_experiments_1996}
Y.~Freund and R.E. Schapire.
\newblock Experiments with a {New} {Boosting} {Algorithm}.
\newblock In \emph{{Proceedings} of the {Thirteenth} {International}
  {Conference} on {International} {Conference} on {Machine} {Learning}}, San
  Francisco, CA, USA, 1996.

\bibitem[Freund and Schapire(1997)]{freund_decision-theoretic_1997}
Y.~Freund and R.E. Schapire.
\newblock A {Decision}-{Theoretic} {Generalization} of {On}-{Line} {Learning}
  and an {Application} to {Boosting}.
\newblock \emph{Journal of Computer and System Sciences}, 55\penalty0
  (1):\penalty0 119--139, 1997.

\bibitem[Friedman(2001)]{friedman_greedy_2001}
J.~Friedman.
\newblock Greedy function approximation: {A} gradient boosting machine.
\newblock \emph{The Annals of Statistics}, 29\penalty0 (5):\penalty0
  1189--1232, 2001.

\bibitem[Friedman(2002)]{friedman_stochastic_2002}
J.~Friedman.
\newblock Stochastic gradient boosting.
\newblock \emph{Computational Statistics \& Data Analysis}, 38\penalty0
  (4):\penalty0 367--378, February 2002.

\bibitem[Friedman et~al.(2000)Friedman, Hastie, and
  Tibshirani]{friedman_additive_2000}
J.~Friedman, T.~Hastie, and R.~Tibshirani.
\newblock Additive logistic regression: a statistical view of boosting (with
  discussion and a rejoinder by the authors).
\newblock \emph{The Annals of Statistics}, 28\penalty0 (2):\penalty0 337--407,
  2000.

\bibitem[Gao and Koller(2011)]{gao_multiclass_2011}
T.~Gao and D.~Koller.
\newblock Multiclass boosting with hinge loss based on output coding.
\newblock In \emph{Proceedings of the 28th {International} {Conference} on
  {International} {Conference} on {Machine} {Learning}}, pages 569--576,
  Madison, WI, USA, 2011. Omnipress.

\bibitem[Grubb and Bagnell(2011)]{grubb_generalized_2011}
A.~Grubb and J.A. Bagnell.
\newblock Generalized {Boosting} {Algorithms} for {Convex} {Optimization}.
\newblock In \emph{{Proceedings} of the 28th {International} {Conference} on
  {Machine} {Learning}}, Bellevue, Washington, USA, 2011.

\bibitem[Hang et~al.(2021)Hang, Huang, Cai, Yang, and Lin]{hang_gradient_2021}
H.~Hang, T.~Huang, Y.~Cai, H.~Yang, and Z.~Lin.
\newblock Gradient {Boosted} {Binary} {Histogram} {Ensemble} for {Large}-scale
  {Regression}.
\newblock \emph{arXiv:2106.01986 [cs, stat]}, 2021.

\bibitem[Ikeagwuani et~al.(2021)Ikeagwuani, Nwonu, and
  Nweke]{ikeagwuani_resilient_2021}
C.~C. Ikeagwuani, D.~C. Nwonu, and C.~C. Nweke.
\newblock Resilient modulus descriptive analysis and estimation for
  fine-grained soils using multivariate and machine learning methods.
\newblock \emph{International Journal of Pavement Engineering}, pages 1--16,
  2021.

\bibitem[Jaggi(2013)]{jaggi_revisiting_2013}
M.~Jaggi.
\newblock Revisiting {Frank}-{Wolfe}: {Projection}-{Free} {Sparse} {Convex}
  {Optimization}.
\newblock In \emph{{Proceedings} of the 30th {International} {Conference} on
  {Machine} {Learning}}, pages 427--435, Atlanta, GA, USA, 2013.

\bibitem[Lin et~al.(2016)Lin, Rosasco, and Zhou]{lin_iterative_2016}
J.~Lin, L.~Rosasco, and D.-X. Zhou.
\newblock Iterative {Regularization} for {Learning} with {Convex} {Loss}
  {Functions}.
\newblock \emph{Journal of Machine Learning Research}, 17\penalty0
  (77):\penalty0 1--38, 2016.

\bibitem[Lin et~al.(2019)Lin, Lei, and Zhou]{lin_boosted_2019}
S.-B. Lin, Y.~Lei, and D.-X. Zhou.
\newblock Boosted {Kernel} {Ridge} {Regression}: {Optimal} {Learning} {Rates}
  and {Early} {Stopping}.
\newblock \emph{Journal of Machine Learning Research}, 20\penalty0
  (46):\penalty0 1--36, 2019.

\bibitem[Lu et~al.(2020)Lu, Karimireddy, Ponomareva, and
  Mirrokni]{lu_accelerating_2020}
H.~Lu, S.~P. Karimireddy, N.~Ponomareva, and V.~Mirrokni.
\newblock Accelerating {Gradient} {Boosting} {Machines}.
\newblock In \emph{Proceedings of the {Twenty} {Third} {International}
  {Conference} on {Artificial} {Intelligence} and {Statistics}}, pages
  516--526, Online, 2020. PMLR.

\bibitem[Mason et~al.(2000{\natexlab{a}})Mason, Baxter, Bartlett, and
  Frean]{mason_boosting_2000}
L.~Mason, J.~Baxter, P.L. Bartlett, and M.~Frean.
\newblock Boosting {Algorithms} as {Gradient} {Descent}.
\newblock In S.A. Solla, T.K. Leen, and K.~Müller, editors, \emph{Advances in
  {Neural} {Information} {Processing} {Systems}}, pages 512--518. MIT Press,
  2000{\natexlab{a}}.

\bibitem[Mason et~al.(2000{\natexlab{b}})Mason, Baxter, Bartlett, and
  Frean]{mason_functional_2000}
L.~Mason, J.~Baxter, P.L. Bartlett, and M.~Frean.
\newblock Functional gradient techniques for combining hypotheses.
\newblock In A.J. Smola, P.L. Bartlett, B.~Shölkopf, and D.~Schuurmans,
  editors, \emph{Advances in {Large} {Margin} {Classifiers}}, pages 221--246.
  The MIT Press, 2000{\natexlab{b}}.

\bibitem[Meir and Rätsch(2003)]{meir_introduction_2003}
R.~Meir and G.~Rätsch.
\newblock An {Introduction} to {Boosting} and {Leveraging}.
\newblock In \emph{Advanced {Lectures} on {Machine} {Learning}}, Lecture
  {Notes} in {Computer} {Science}, pages 118--183. Springer, Berlin,
  Heidelberg, 2003.

\bibitem[Nesterov(1983)]{nesterov_method_1983}
Y.~Nesterov.
\newblock A method of solving a convex programming problem with convergence
  rate \({O}(1/k^2)\).
\newblock \emph{Soviet Mathematics Doklady}, 27, 1983.

\bibitem[Nesterov(2004)]{nesterov_introductory_2004}
Y.~Nesterov.
\newblock \emph{Introductory {Lectures} on {Convex} {Optimization}: {A} {Basic}
  {Course}}.
\newblock Kluwer Academic Publishers, 2004.

\bibitem[Ochs et~al.(2014)Ochs, Chen, Brox, and Pock]{ochs_ipiano:_2014}
P.~Ochs, Y.~Chen, T.~Brox, and T.~Pock.
\newblock {iPiano}: {Inertial} {Proximal} {Algorithm} for {Nonconvex}
  {Optimization}.
\newblock \emph{SIAM {J}ournal on {I}maging {S}ciences}, 2014.

\bibitem[Park et~al.(2009)Park, Lee, and Ha]{park_l_2_2009}
B.~U. Park, Y.~K. Lee, and S.~Ha.
\newblock \({L}_2\) boosting in kernel regression.
\newblock \emph{Bernoulli}, 15\penalty0 (3):\penalty0 599--613, 2009.

\bibitem[Pedregosa et~al.(2011)Pedregosa, Varoquaux, Gramfort, Michel, Thirion,
  Grisel, Blondel, Prettenhofer, Weiss, Dubourg, Vanderplas, Passos,
  Cournapeau, Brucher, Perrot, and Duchesnay]{pedregosa_scikit-learn:_2011}
F.~Pedregosa, G.~Varoquaux, A.~Gramfort, V.~Michel, B.~Thirion, O.~Grisel,
  M.~Blondel, P.~Prettenhofer, R.~Weiss, V.~Dubourg, J.~Vanderplas, A.~Passos,
  D.~Cournapeau, M.~Brucher, M.~Perrot, and E.~Duchesnay.
\newblock Scikit-learn: {Machine} {Learning} in {Python}.
\newblock \emph{Journal of Machine Learning Research}, 12:\penalty0 2825--2830,
  2011.

\bibitem[Rajendran et~al.(2021)Rajendran, Srinivas, and
  Grimshaw]{rajendran_predicting_2021}
S.~Rajendran, S.~Srinivas, and T.~Grimshaw.
\newblock Predicting demand for air taxi urban aviation services using machine
  learning algorithms.
\newblock \emph{Journal of Air Transport Management}, 92:\penalty0 102043,
  2021.

\bibitem[Rockafellar(1976)]{rockafellar1976monotone}
R~Tyrrell Rockafellar.
\newblock Monotone operators and the proximal point algorithm.
\newblock \emph{SIAM {J}ournal on {C}ontrol and {O}ptimization}, 14\penalty0
  (5):\penalty0 877--898, 1976.

\bibitem[Rätsch et~al.(2002)Rätsch, Mika, and
  Warmuth]{ratsch_convergence_2002}
G.~Rätsch, S.~Mika, and M.K. Warmuth.
\newblock On the {Convergence} of {Leveraging}.
\newblock In T.G. Dietterich, S.~Becker, and Z.~Ghahramani, editors,
  \emph{Advances in {Neural} {Information} {Processing} {Systems}}, pages
  487--494. MIT Press, 2002.

\bibitem[Santana et~al.(2021)Santana, Silveira, Sobrinho, Silva, Silva, Santos,
  Gurjão, and Perkusich]{santana_classification_2021}
I.~V. D.~S. Santana, A.~C.~M. Silveira, A.~Sobrinho, L.~C. Silva, L.~D. Silva,
  D.~F.~S. Santos, E.~C. Gurjão, and A.~Perkusich.
\newblock Classification {Models} for {COVID}-19 {Test} {Prioritization} in
  {Brazil}: {Machine} {Learning} {Approach}.
\newblock \emph{Journal of Medical Internet Research}, 23\penalty0
  (4):\penalty0 e27293, 2021.

\bibitem[Schapire(1990)]{schapire_strength_1990}
R.E. Schapire.
\newblock The strength of weak learnability.
\newblock \emph{Machine Learning}, 5\penalty0 (2):\penalty0 197--227, 1990.

\bibitem[Temlyakov(2014)]{temlyakov_greedy_2014}
V.~N. Temlyakov.
\newblock Greedy expansions in convex optimization.
\newblock \emph{{Proceedings} of the {Steklov} {Institute} of {Mathematics}},
  284:\penalty0 244--262, 2014.

\bibitem[Tyralis and Papacharalampous(2021)]{tyralis_boosting_2021}
H.~Tyralis and G.~Papacharalampous.
\newblock Boosting algorithms in energy research: a systematic review.
\newblock \emph{Neural Computing and Applications}, 33\penalty0 (21):\penalty0
  14101--14117, 2021.

\bibitem[Wang et~al.(2015)Wang, Wang, E, and Schapire]{wang_functional_2015}
C.~Wang, Y.~Wang, W.~E, and R.~Schapire.
\newblock Functional {Frank}-{Wolfe} {Boosting} for {General} {Loss}
  {Functions}.
\newblock \emph{arXiv:1510.02558 [cs, stat]}, 2015.

\bibitem[Wang et~al.(2019)Wang, Liao, and Lin]{wang_rescaled_2019}
Y.~Wang, X.~Liao, and S.~Lin.
\newblock Rescaled {Boosting} in {Classification}.
\newblock \emph{IEEE Transactions on Neural Networks and Learning Systems},
  30\penalty0 (9):\penalty0 2598--2610, 2019.

\bibitem[Zeng et~al.(2022)Zeng, Zhang, and Lin]{zeng_fully_2022}
J.~Zeng, M.~Zhang, and S.-B. Lin.
\newblock Fully corrective gradient boosting with squared hinge: {Fast}
  learning rates and early stopping.
\newblock \emph{Neural Networks}, 147:\penalty0 136--151, 2022.

\bibitem[Zhang(2002)]{zhang_general_2002}
T.~Zhang.
\newblock A {General} {Greedy} {Approximation} {Algorithm} with {Applications}.
\newblock In T.G. Dietterich, S.~Becker, and Z.~Ghahramani, editors,
  \emph{Advances in {Neural} {Information} {Processing} {Systems}}, pages
  1065--1072. MIT Press, 2002.

\bibitem[Zhang(2003)]{zhang_sequential_2003}
T.~Zhang.
\newblock Sequential greedy approximation for certain convex optimization
  problems.
\newblock \emph{IEEE Transactions on Information Theory}, 49\penalty0
  (3):\penalty0 682--691, March 2003.

\bibitem[Zhang and Yu(2005)]{zhang_boosting_2005}
T.~Zhang and B.~Yu.
\newblock Boosting with early stopping: {Convergence} and consistency.
\newblock \emph{The Annals of Statistics}, 33\penalty0 (4):\penalty0
  1538--1579, 2005.

\end{thebibliography}

	\appendix
	\section{Analysis of the approximated proximal point method}
		\label{app:proof_cv}
\subsection{Setting}
  Let us consider the optimization problem
  \begin{opb}{opb:app_prox}
    \minimize{x \in \R^n} & F(x),
  \end{opb}
  where $F : \R^n \to \R$ is convex.

  For an operator $P : \R^n \to \R^n$, we consider the approximated proximal point method, described in Algorithm~\ref{alg:app_prox}, as well as the approximated proximal point method with accumulation, described in Algorithm~\ref{alg:app_prox_acc}.
  Both are similar to the proximal point iteration but makes use of a modified direction of update ($P(g_t)$ or \(P(g_t+\Delta_t)\) instead of $g_t$).
  In particular, let us remark that when $P(x) = x$, Algorithms~\ref{alg:app_prox} and \ref{alg:app_prox_acc} recover the original proximal point method.
  \begin{algorithm}[ht]
     \begin{algorithmic}[1]
       \REQUIRE
       \(T\) (number of iterations),
       $\lambda_0, \dots, \lambda_{T-1}>0$ (proximal steps),
       $P : \R^n \to \R^n$ (approximation operator).
       \STATE Set $x_0 \in \R^n$
       \COMMENT{initialization}.
       \FOR{$t=0$ \TO $T-1$}
         \STATE	$g_t \gets \frac{1}{\lambda_t} \left( x_t - \prox_{\lambda_t F} (x_t) \right)$.
         \STATE $x_{t+1} \gets x_t - \lambda_t P(g_t)$.
       \ENDFOR
       \ENSURE $x_T$.
     \end{algorithmic}
     \caption{Approximated proximal point method.}
     \label{alg:app_prox}
  \end{algorithm}

  \begin{algorithm}[ht]
     \begin{algorithmic}[1]
       \REQUIRE
       \(T\) (number of iterations),
       $\lambda_0, \dots, \lambda_{T-1}>0$ (proximal steps),
       $P : \R^n \to \R^n$ (approximation operator).
       \STATE Set $x_0 \in \R^n$ and $\Delta_0 = 0$
       \COMMENT{initialization}.
       \FOR{$t=0$ \TO $T-1$}
         \STATE	$g_t \gets \frac{1}{\lambda_t} \left( x_t - \prox_{\lambda_t F} (x_t) \right)$.
         \STATE $x_{t+1} \gets x_t - \lambda_t P(g_t + \Delta_t)$.
         \STATE \(\Delta_{t+1} = g_t + \Delta_t - P(g_t + \Delta_t)\).
       \ENDFOR
       \ENSURE $x_T$.
     \end{algorithmic}
     \caption{Approximated proximal point method with accumulation.}
     \label{alg:app_prox_acc}
  \end{algorithm}

  The forthcoming sections prove convergence of Algorithm~\ref{alg:app_prox} for strongly convex functions with Lipschitz continuous gradient (linear rate exhibited in Theorem~\ref{thm:cv_app_prox_sc_sm}) and of Algorithm~\ref{alg:app_prox_acc} for Lipschitz continuous functions (sublinear rate exhibited in Theorem~\ref{thm:cv_app_prox_l}).
  To be more formal, the following assumptions will be used:
  \begin{assumptions}
    \hyp{SM} \(F\) is $L$-smooth (for some $L>0$): \(F\) is differentiable and
    \[
      \forall x, x' \in \R^n,
      \qquad
    	F(x') \le F(x) + \iprod{\nabla F(x)}{x'-x} + \frac{L}{2} \norm{x'-x}^2.
    \] \label{hyp:str_smooth}
    \hyp{SC} \(F\) is $\kappa$-strongly convex (for some $\kappa>0$):
    \[
      \forall x, x' \in \R^n,
      \forall \eta \in \partial F(x),
      \qquad
      F(x') \ge F(x) + \iprod{\eta}{x'-x} + \frac{\kappa}{2} \norm{x'-x}^2.
    \] \label{hyp:str_cvx}
    \hyp{L} \(F\) is $G$-Lipschitz continuous (for some $G>0$):
    \[
      \forall x \in \R^n, \forall \eta \in \partial F(x),
      \qquad
      \norm{\eta} \le G.
    \] \label{hyp:lipf}
  \end{assumptions}

  In any case, it is assumed that:
  \begin{assumptions}
    \hyp{E} There exists $\zeta \in (0, 1]$ such that for all \(g \in \R^n\), \(\norm{g - P(g)}^2 \le (1-\zeta^2) \norm{g}^2\). \label{hyp:edge}
  \end{assumptions}

  Assumption~\ref{hyp:edge} is often referred to as the edge property and is quite standard in the literature \citep{grubb_generalized_2011}.
  It measures the error of the approximated operator \(P\) on the direction of descent \(g_t\).

\subsection{Strongly convex function with smooth gradient}
  \begin{theorem}
  	Let $(x_t)_{t}$ be a sequence generated by Algorithm~\ref{alg:app_prox}.
    Assume that Assumptions~\ref{hyp:edge}, \ref{hyp:str_smooth} and \ref{hyp:str_cvx} hold.
    Let $\{x^\star\} = \argmin_{x \in \R^n} F(x)$ (well defined by strong convexity), and choose $\lambda_t = \frac{\zeta^2}{8L}$.
  	Then,
  	\[
  		F(x_T) - F(x^\star) \leq
      \left( 1- \frac{\zeta^4  \kappa}{21L} \right)^T
      \left( F(x_0) - F(x^\star) \right).
  	\]
    \label{thm:cv_app_prox_sc_sm}
  \end{theorem}

  \begin{proof}
    First of all, let us remark that:
    \begin{enumerate}
      \item Assumption~\ref{hyp:str_smooth} implies $L$-Lipschitz continuity of the gradient $\nabla F$ \cite[Theorem~2.1.5]{nesterov_introductory_2004}:
      \begin{equation}
        \forall x, x' \in \R^n, \qquad
        \norm{\nabla F(x) - \nabla F(x')} \le L \norm{x-x'};
        \label{equ:lipF}
      \end{equation}
        \item Assumption~\ref{hyp:str_cvx} leads to the upper bound \cite[Theorem~2.1.10]{nesterov_introductory_2004}:
        \begin{equation}
          \forall x \in \R^n, \qquad
          2\kappa \left( F(x) - F(x^\star) \right) \leq \norm{\nabla F(x)}^2.
          \label{equ:control_kappa}
        \end{equation}
    \end{enumerate}

    Then, from Assumption~\ref{hyp:str_smooth} and by the update rule for $x_{t+1}$ in Algorithm~\ref{alg:app_prox}:
    \begin{align*}
      F(x_{t+1})
      &\leq F(x_t) + \langle \nabla F(x_t) , -\lambda_t P(g_t) \rangle + \frac{L \lambda_t^2 }{2}\norm{P(g_t)}^2 \\
      &= F(x_t) -\lambda_t \langle g_t ,  P(g_t) \rangle  -\lambda_t \langle \nabla F(x_t) - g_t ,  P(g_t) \rangle + \frac{L \lambda_t^2 }{2} \norm{P(g_t)}^2.
      \numberthis \label{equ:proof1}
    \end{align*}

    Now, from Assumption~\ref{hyp:edge}:
   	\begin{align*}
    	-\lambda_t \langle g_t ,  P(g_t) \rangle
    	&= \frac{\lambda_t}{2} \left( \norm{g_t - P(g_t)}^2 - \norm{g_t}^2- \norm{ P(g_t)}^2 \right) \\
    	&\leq\frac{\lambda_t}{2} \left[ (1-\zeta^2) \norm{g_t}^2 - \norm{g_t}^2- \norm{ P(g_t)}^2 \right] \\
    	&= -\frac{\lambda_t\zeta^2}{2} \norm{g_t}^2 - \frac{\lambda_t}{2}\norm{ P(g_t)}^2.
      \numberthis \label{equ:proof2}
   	\end{align*}

    Besides, given that $g_t = \nabla F(\prox_{\lambda_t F}(x_t))$ by definition of the proximal operator, one has:
    \begin{align*}
      \norm{\nabla F(x_t) - g_t}
      &= \norm{\nabla F(x_t) - \nabla F\left( \prox_{\lambda_t F}(x_t) \right)} \\
      &\leq  L \norm{x_t - \prox_{\lambda_t F}(x_t)}
      & (\text{Equation~\eqref{equ:lipF}})\\
      &\leq \lambda_t L \norm{g_t}
      & (\text{definition of } g_t).
      \numberthis \label{equ:grad_prox}
   	\end{align*}

    So,
   	\begin{align*}
      -\lambda_t \langle \nabla F(x_t) - g_t ,  P(g_t) \rangle
      &\leq \lambda_t \norm{\nabla F(x_t) - g_t} \norm{P(g_t)}
      & (\text{Cauchy-Schwarz}) \\
      & \leq \lambda_t^2 L \norm{g_t} \norm{P(g_t)}
      & (\text{Equation~\eqref{equ:grad_prox}}) \\
      & \leq 2\lambda_t^2 L \norm{g_t}^2,
      \numberthis \label{equ:proof3}
   	\end{align*}
    since \(\norm{P(g_t)} \le 2 \norm{g_t}\), by Assumption~\ref{hyp:edge}.

    Combining Equations~\eqref{equ:proof1}, \eqref{equ:proof2} and \eqref{equ:proof3}:
   	\begin{align*}
    	F(x_{t+1})
      &\leq F(x_t) - \frac{\lambda_t \zeta^2 }{2} \norm{g_t}^2 - \frac{\lambda_t}{2}\norm{ P(g_t)}^2 + 2\lambda_t^2 L \norm{g_t}^2 + \frac{L \lambda_t^2}{2} \norm{P(g_t)}^2 \\
    	&= F(x_t) - \lambda_t \left( \frac{\zeta^2 }{2}-2\lambda_t L \right)\norm{g_t}^2 - \frac{\lambda_t}{2} \left( 1 -  L\lambda_t \right) \norm{ P(g_t)}^2.
   	\end{align*}
    Now, choosing $\lambda_t=\frac{\zeta^2}{8L}$, one has
    $\lambda_t \left( \frac{\zeta^2 }{2}-2\lambda_t L \right) = \frac{\zeta^4}{32L}$ on one hand and
    $- \frac{\lambda_t}{2} \left( 1 -  L\lambda_t \right) \norm{ P(g_t)}^2 \le 0$ on the other, leading to:
   	\begin{align}
    	F(x_{t+1})
      &\leq F(x_t) -\frac{\zeta^4 }{32L} \norm{g_t}^2.
      \label{equ:proof4}
   	\end{align}

    Let us remark that, by Equation~\eqref{equ:control_kappa}:
    \begin{align*}
      2 \kappa \left( F(x_t) - F(x^\star) \right)
      &\leq \norm{\nabla F(x_t)}^2 \\
      &\leq \left( \norm{\nabla F(x_t) - g_t} + \norm{g_t} \right)^2 \\
      &\leq (1+\lambda_t L)^2 \norm{g_t}^2
      & (\text{Equation~\eqref{equ:grad_prox}}) \\
      &\leq \left( 1+\frac{\zeta^2}{8} \right)^2 \norm{g_t}^2
      & \left(\lambda_t=\frac{\zeta^2}{8L}\right),
    \end{align*}
    that is,
    \begin{align*}
      \norm{g_t}^2
      &\ge \frac{128 \kappa}{(8+\zeta^2)^2} \left( F(x_t) - F(x^\star) \right).
    \end{align*}

    So, from Equation~\eqref{equ:proof4},
   	\begin{align*}
   		F(x_{t+1}) -F(x^\star)
   		&\leq F(x_{t}) -F(x^\star) -\frac{\zeta^4 }{32L} \norm{g_t}^2 \\
   		&\leq \left( 1- \frac{\zeta^4}{32L} \frac{128 \kappa}{(8 + \zeta^2)^2} \right) \left( F(x_{t}) -F(x^\star) \right) \\
   		&= \left( 1- \frac{4 \zeta^4}{(8 + \zeta^2)^2} \frac{\kappa}{L} \right) \left( F(x_{t}) -F(x^\star) \right) \\
   		&\le \left( 1- \frac{\zeta^4}{21} \frac{\kappa}{L} \right) \left( F(x_{t}) -F(x^\star) \right)
      & (\star) \\
      &\le \left( 1- \frac{\zeta^4 \kappa}{21L} \right)^{t+1} \left( F(x_{0}) -F(x^\star) \right)
      & (\text{by induction}),
   	\end{align*}
    where we have used \((\star)\) that \(\forall x \in [0, 1], \frac{4 x^2}{(8+x)^2} \ge \frac{x^2}{21}\).
   \end{proof}

\subsection{Lipschitz continuous convex function}

\begin{lemma}
  Let $(x_t)_{t}$ be a sequence generated by Algorithm~\ref{alg:app_prox_acc}.
  Assume that Assumptions~\ref{hyp:edge}, \ref{hyp:str_cvx} and \ref{hyp:lipf} hold and that there exists $x^\star \in \argmin_{x \in \R^n} F(x)$.
  Then,
  \begin{align*}
    &\min_{1 \le t \le T} F(x_t) - F(x^\star) \\
    &\le \frac{1}{2T} \left( \frac{1}{\lambda_0} - \kappa \right) \norm{x_0 - x^\star}^2
    + \frac{1}{2T} \sum_{t=1}^{T-1} \left( \frac{1}{\lambda_t} - \frac{1}{\lambda_{t-1}} - \kappa \right) \norm{x_t - x^\star}^2 \\
    &+ \frac 1T \sum_{t=0}^{T-1} \lambda_t \bigg( \frac{1}{2} \norm{P(g_t+\Delta_t)}^2 - \left( 1 + \frac{\kappa \lambda_t}{2} \right) \norm{g_t}^2
    + \kappa \iprod{g_t}{x_t-x^\star} \\
    &+ \iprod{\Delta_{t+1}}{P(g_t + \Delta_t)}
    + G \norm{g_t - P(g_t + \Delta_t)} \bigg) + \frac{\lambda_{T-1}}{2T} \norm{\Delta_{T}}^2.
  \end{align*}
  In addition, the result still holds if \(\kappa = 0\).
  \label{lem:cv_inequality}
\end{lemma}

\begin{proof}[Lemma~\ref{lem:cv_inequality}]
  For any non-negative integer \(t < T\), let
  \[
    y_{t+1} = x_t - \lambda_t g_t = \prox_{\lambda_t F}(x_t).
  \]
  By construction, \(g_t \in \partial F(y_{t+1})\), so
  \begin{align}
    F(x^\star)
    &\ge F(y_{t+1}) + \iprod{g_t}{x^\star - y_{t+1}} + \frac{\kappa}{2} \norm{y_{t+1} - x^\star}^2 \notag \\
    &= F(y_{t+1}) + \iprod{g_t}{x^\star - (x_t - \lambda_t g_t)} + \frac{\kappa}{2} \norm{(x_t - \lambda_t g_t) - x^\star}^2 \notag \\
    &= F(y_{t+1}) + \iprod{g_t}{x^\star - x_t} + \lambda_t \norm{g_t}^2 + \frac{\kappa}{2} \norm{x_t - x^\star}^2 + \frac{\kappa \lambda_t^2}{2} \norm{g_t}^2 \notag \\
    &- \kappa \lambda_t \iprod{g_t}{x_t-x^\star} \notag \\
    &= F(y_{t+1}) + \iprod{P(g_t+\Delta_t)}{x^\star - x_t} + \left( \lambda_t + \frac{\kappa \lambda_t^2}{2} \right) \norm{g_t}^2 + \frac{\kappa}{2} \norm{x_t - x^\star}^2 \notag \\
    &- \kappa \lambda_t \iprod{g_t}{x_t-x^\star} + \iprod{g_t - P(g_t+\Delta_t)}{x^\star - x_t}.
    \label{equ:conv_xstar}
  \end{align}

  Now, let us analyze the potential \(\norm{x_{t+1} - x^\star}^2\):
  \begin{align*}
    \norm{x_{t+1} - x^\star}^2
    &= \norm{x_t - \lambda_t P(g_t+\Delta_t) - x^\star}^2 \\
    &= \norm{x_t - x^\star}^2 + \lambda_t^2 \norm{P(g_t+\Delta_t)}^2 - 2\lambda_t \iprod{P(g_t+\Delta_t)}{x_t-x^\star}.
  \end{align*}
  Thus,
  \begin{align}
    \iprod{P(g_t+\Delta_t)}{x_t-x^\star}
    &= \frac{1}{2\lambda_t} \left( \norm{x_t - x^\star}^2 - \norm{x_{t+1} - x^\star}^2 \right) + \frac{\lambda_t}{2} \norm{P(g_t+\Delta_t)}^2.
    \label{equ:potential}
  \end{align}

  Combining Equation~\eqref{equ:conv_xstar} and Equation~\eqref{equ:potential}, we obtain:
  \begin{align}
    F(y_{t+1}) - F(x^\star)
    &\le \iprod{P(g_t+\Delta_t)}{x_t-x^\star} - \left( \lambda_t + \frac{\kappa \lambda_t^2}{2} \right) \norm{g_t}^2 - \frac{\kappa}{2} \norm{x_t - x^\star}^2 \notag \\
    &+ \kappa \lambda_t \iprod{g_t}{x_t-x^\star} + \iprod{g_t - P(g_t+\Delta_t)}{x_t-x^\star} \notag \\
    &\le \frac{1}{2\lambda_t} \left( \norm{x_t - x^\star}^2 - \norm{x_{t+1} - x^\star}^2 \right) - \frac{\kappa}{2} \norm{x_t - x^\star}^2 \notag \\
    &+ \frac{\lambda_t}{2} \norm{P(g_t+\Delta_t)}^2 - \left( \lambda_t + \frac{\kappa \lambda_t^2}{2} \right) \norm{g_t}^2 \notag \\
    &+ \kappa \lambda_t \iprod{g_t}{x_t-x^\star} + \iprod{g_t - P(g_t+\Delta_t)}{x_t-x^\star}.
    \label{equ:diff_F}
  \end{align}

  Now, remark that:
  \begin{align}
    & \sum_{t=0}^{T-1} \left( \left(\frac{1}{\lambda_t} - \kappa \right) \norm{x_t - x^\star}^2 - \frac{1}{\lambda_t} \norm{x_{t+1} - x^\star}^2 \right) \notag \\
    &= \left( \frac{1}{\lambda_0} - \kappa \right) \norm{x_0 - x^\star}^2 + \sum_{t=1}^{T-1} \left( \frac{1}{\lambda_t} - \kappa \right) \norm{x_t - x^\star}^2
    - \sum_{t=0}^{T-2} \frac{1}{\lambda_t} \norm{x_{t+1} - x^\star}^2 \notag \\
    &- \frac{1}{\lambda_{T-1}} \norm{x_T - x^\star}^2 \notag \\
    &= \left( \frac{1}{\lambda_0} - \kappa \right) \norm{x_0 - x^\star}^2
    + \sum_{t=1}^{T-1} \left( \frac{1}{\lambda_t} - \frac{1}{\lambda_{t-1}} - \kappa \right) \norm{x_t - x^\star}^2 \notag \\
    &- \frac{1}{\lambda_{T-1}} \norm{x_T - x^\star}^2,
    \label{equ:telescopic_sum}
  \end{align}
  and
  \begin{align}
    & \sum_{t=0}^{T-1} \iprod{g_t - P(g_t+\Delta_t)}{x_t-x^\star} \notag \\
    &= \sum_{t=0}^{T-1} \iprod{g_t + \Delta_t - P(g_t+\Delta_t)}{x_{t+1} + \lambda_t P(g_t + \Delta_t) - x^\star} - \sum_{t=0}^{T-1} \iprod{\Delta_t}{x_t-x^\star} \notag \\
    &= \sum_{t=0}^{T-1} \iprod{\Delta_{t+1}}{x_{t+1} - x^\star} - \sum_{t=0}^{T-1} \iprod{\Delta_t}{x_t-x^\star}
    + \sum_{t=0}^{T-1} \lambda_t \iprod{\Delta_{t+1}}{P(g_t + \Delta_t)} \notag \\
    &= \iprod{\Delta_{T}}{x_{T} - x^\star} - \iprod{\Delta_0}{x_0-x^\star}
    + \sum_{t=0}^{T-1} \lambda_t \iprod{\Delta_{t+1}}{P(g_t + \Delta_t)} \notag \\
    &= \iprod{\Delta_{T}}{x_{T} - x^\star} + \sum_{t=0}^{T-1} \lambda_t \iprod{\Delta_{t+1}}{P(g_t + \Delta_t)},
    \label{equ:telescopic_delta}
  \end{align}
  since \(\Delta_0 = 0\).

  Then, by summation of Equation~\eqref{equ:diff_F} and using Equation~\eqref{equ:telescopic_sum},
  \begin{align*}
    &\sum_{t=0}^{T-1} (F(y_{t+1}) - F(x^\star)) \\
    &\le \frac{1}{2} \sum_{t=0}^{T-1} \left( \left(\frac{1}{\lambda_t} - \kappa \right) \norm{x_t - x^\star}^2 - \frac{1}{\lambda_t} \norm{x_{t+1} - x^\star}^2 \right) \\
    &+ \sum_{t=0}^{T-1} \lambda_t \left( \frac{1}{2} \norm{P(g_t+\Delta_t)}^2 - \left( 1 + \frac{\kappa \lambda_t}{2} \right) \norm{g_t}^2 + \kappa \iprod{g_t}{x_t-x^\star} \right) \\
    & + \sum_{t=0}^{T-1} \iprod{g_t - P(g_t+\Delta_t)}{x_t-x^\star} \\
    &= \left( \frac{1}{2\lambda_0} - \frac{\kappa}{2} \right) \norm{x_0 - x^\star}^2
    + \frac{1}{2} \sum_{t=1}^{T-1} \left( \frac{1}{\lambda_t} - \frac{1}{\lambda_{t-1}} - \kappa \right) \norm{x_t - x^\star}^2 \\
    &+ \sum_{t=0}^{T-1} \lambda_t \left( \frac{1}{2} \norm{P(g_t+\Delta_t)}^2 - \left( 1 + \frac{\kappa \lambda_t}{2} \right) \norm{g_t}^2 + \kappa \iprod{g_t}{x_t-x^\star} \right) \\
    & + \sum_{t=0}^{T-1} \iprod{g_t - P(g_t+\Delta_t)}{x_t-x^\star} - \frac{1}{2\lambda_{T-1}} \norm{x_T - x^\star}^2.
  \end{align*}

  Now, using Equation~\eqref{equ:telescopic_delta},
  \begin{align*}
    &\sum_{t=0}^{T-1} (F(y_{t+1}) - F(x^\star))\\
    &\le \left( \frac{1}{2\lambda_0} - \frac{\kappa}{2} \right) \norm{x_0 - x^\star}^2
    + \frac{1}{2} \sum_{t=1}^{T-1} \left( \frac{1}{\lambda_t} - \frac{1}{\lambda_{t-1}} - \kappa \right) \norm{x_t - x^\star}^2 \\
    &+ \sum_{t=0}^{T-1} \lambda_t \bigg( \frac{1}{2} \norm{P(g_t+\Delta_t)}^2 - \left( 1 + \frac{\kappa \lambda_t}{2} \right) \norm{g_t}^2 + \kappa \iprod{g_t}{x_t-x^\star} \\
    &+ \iprod{\Delta_{t+1}}{P(g_t + \Delta_t)} \bigg) + \iprod{\Delta_{T}}{x_{T} - x^\star} - \frac{1}{2\lambda_{T-1}} \norm{x_T - x^\star}^2 \\
    &\le \left( \frac{1}{2\lambda_0} - \frac{\kappa}{2} \right) \norm{x_0 - x^\star}^2
    + \frac{1}{2} \sum_{t=1}^{T-1} \left( \frac{1}{\lambda_t} - \frac{1}{\lambda_{t-1}} - \kappa \right) \norm{x_t - x^\star}^2 \\
    &+ \sum_{t=0}^{T-1} \lambda_t \bigg( \frac{1}{2} \norm{P(g_t+\Delta_t)}^2 - \left( 1 + \frac{\kappa \lambda_t}{2} \right) \norm{g_t}^2 + \kappa \iprod{g_t}{x_t-x^\star} \\
    &+ \iprod{\Delta_{t+1}}{P(g_t + \Delta_t)} \bigg) + \frac{\lambda_{T-1}}{2} \norm{\Delta_{T}}^2,
  \end{align*}
  where the last line comes from \(bx - ax^2 \le \frac{b^2}{4a}\) for any \(a > 0\) and \(b \in \R\).

  To conclude,
  \begin{align*}
    &\min_{1 \le t \le T} F(x_t) - F(x^\star) \\
    &\le \frac 1T \sum_{t=0}^{T-1} \left(F(y_{t+1}) - F(x^\star) \right) + \frac 1T \sum_{t=0}^{T-1} \left(F(x_{t+1}) - F(y_{t+1}) \right) \\
    & \le \frac 1T \sum_{t=0}^{T-1} \left(F(y_{t+1}) - F(x^\star) \right) + \frac 1T \sum_{t=0}^{T-1} G \norm{(x_{t} - \lambda_t P(g_t + \Delta_t)) - (x_t - \lambda_t g_t)} \\
    & \le \frac{1}{2T} \left( \frac{1}{\lambda_0} - \kappa \right) \norm{x_0 - x^\star}^2
    + \frac{1}{2T} \sum_{t=1}^{T-1} \left( \frac{1}{\lambda_t} - \frac{1}{\lambda_{t-1}} - \kappa \right) \norm{x_t - x^\star}^2 \\
    &+ \frac 1T \sum_{t=0}^{T-1} \lambda_t \bigg( \frac{1}{2} \norm{P(g_t+\Delta_t)}^2 - \left( 1 + \frac{\kappa \lambda_t}{2} \right) \norm{g_t}^2 + \kappa \iprod{g_t}{x_t-x^\star} \\
    &+ \iprod{\Delta_{t+1}}{P(g_t + \Delta_t)} + G \norm{g_t - P(g_t + \Delta_t)} \bigg) + \frac{\lambda_{T-1}}{2T} \norm{\Delta_{T}}^2.
  \end{align*}
\end{proof}

  \begin{theorem}
  	Let $(x_t)_{t}$ be a sequence generated by Algorithm~\ref{alg:app_prox_acc}.
    Assume that Assumptions~\ref{hyp:edge} and \ref{hyp:lipf} hold.
    Assume also that there exists a minimizer $x^\star \in \argmin_{x \in \R^n} F(x)$ and that \(\norm{x_t} \le R\) and \(\norm{x^\star} \le R\) (for some \(R>0\) and all \(t\)).
    Then, choosing $\lambda_t = \frac{1}{\sqrt{t+1}}$ leads to:
    \begin{align*}
      \min_{1 \le t \le T} F(x_t) - F(x^\star)
      &\le \frac{2R^2}{\sqrt T} + \frac{2G^2}{\zeta^4 \sqrt T} \left( 20 + \frac{1}{T} \right).
    \end{align*}
    \label{thm:cv_app_prox_l}
  \end{theorem}

  \begin{proof}
    By Lemma~\ref{lem:cv_inequality} with \(\kappa=0\) and \(\lambda_t = \frac{1}{\sqrt{t+1}}\), we have:
    \begin{align*}
      \min_{1 \le t \le T} F(x_t) - F(x^\star)
      & \le \frac{1}{2 \lambda_0 T} \norm{x_0 - x^\star}^2
      + \frac{1}{2T} \sum_{t=1}^{T-1} \left( \frac{1}{\lambda_t} - \frac{1}{\lambda_{t-1}} \right) \norm{x_t - x^\star}^2 \\
      &+ \frac 1T \sum_{t=0}^{T-1} \lambda_t \bigg( \frac{1}{2} \norm{P(g_t+\Delta_t)}^2 - \norm{g_t}^2 \\
      &+ \iprod{\Delta_{t+1}}{P(g_t + \Delta_t)} + G \norm{g_t - P(g_t + \Delta_t)} \bigg) \\
      &+ \frac{\lambda_{T-1}}{2T} \norm{\Delta_{T}}^2 \\
      & \le
      \frac{1}{2T} \sum_{t=0}^{T-1} \left( \sqrt{t+1} - \sqrt{t} \right) \norm{x_t - x^\star}^2 \\
      &+ \frac 1T \sum_{t=0}^{T-1} \frac{1}{\sqrt{t+1}} \bigg( \frac{1}{2} \norm{P(g_t+\Delta_t)}^2
      + \iprod{\Delta_{t+1}}{P(g_t + \Delta_t)} \\
      &+ G \norm{g_t - P(g_t + \Delta_t)} \bigg)
      + \frac{1}{2T^{\frac 32}} \norm{\Delta_{T}}^2 \\
      &\le \frac{2R^2}{\sqrt T}
      + \frac 1T \sum_{t=0}^{T-1} \frac{1}{\sqrt{t+1}} \bigg( \frac{1}{2} \norm{P(g_t+\Delta_t)}^2 \\
      &+ \iprod{\Delta_{t+1}}{P(g_t + \Delta_t)}
      + G \norm{g_t - P(g_t + \Delta_t)} \bigg) \\
      &+ \frac{1}{2T^{\frac 32}} \norm{\Delta_{T}}^2.
    \end{align*}

    Now, since \(g_t \in \partial F(y_{t+1})\), \(\norm{g_t} \le G\).
    In addition, since \(\Delta_0 = 0\), by Assumption~\ref{hyp:edge},
    \begin{align*}
      \norm{\Delta_{T+1}}
      &\le \sqrt{1-\zeta^2} \norm{g_T + \Delta_{T}} \\
      &\le \sqrt{1-\zeta^2} \norm{g_T} + \sqrt{1-\zeta^2} \norm{\Delta_{T}} \\
      &\le \sum_{t=0}^T \sqrt{1-\zeta^2}^{T+1-t} \norm{g_t} \\
      &\le G \sum_{t=1}^{T+1} \sqrt{1-\zeta^2}^t \\
      &\le \frac{\sqrt{1-\zeta^2}}{1 - \sqrt{1-\zeta^2}} G \\
      &\le \frac{2}{\zeta^2} G,
    \end{align*}
    where we have used that \(\frac{1}{1 - \sqrt{1-\zeta^2}} \le \frac{2}{\zeta^2}\).
    Moreover,
    \begin{align*}
      \norm{P(g_t + \Delta_t)}
      &\le \norm{P(g_t + \Delta_t) - (g_t+\Delta_t)} + \norm{g_t + \Delta_t} \\
      &\le (\sqrt{1-\zeta^2} + 1) \norm{g_t+\Delta_t} \\
      &\le (\sqrt{1-\zeta^2} + 1) G + (\sqrt{1-\zeta^2} + 1) \frac{\sqrt{1-\zeta^2}}{1 - \sqrt{1-\zeta^2}} G \\
      &\le \frac{(1 - (1-\zeta^2)) + (\sqrt{1-\zeta^2} + 1) \sqrt{1-\zeta^2}}{1 - \sqrt{1-\zeta^2}} G \\
      &\le \frac{\zeta^2 + (1-\zeta^2) + \sqrt{1-\zeta^2}}{1 - \sqrt{1-\zeta^2}} G \\
      &\le \frac{1 + \sqrt{1-\zeta^2}}{1 - \sqrt{1-\zeta^2}} G \\
      &\le \frac{4}{\zeta^2} G.
    \end{align*}
    At last,
    \begin{align*}
      \norm{g_t - P(g_t + \Delta_t)}
      &\le \norm{g_t + \Delta_t - P(g_t + \Delta_t)} + \norm{\Delta_t} \\
      &\le \sqrt{1-\zeta^2} \norm{g_t+\Delta_t} + \norm{\Delta_t} \\
      &\le \sqrt{1-\zeta^2} \norm{g_t} + (\sqrt{1-\zeta^2} + 1) \norm{\Delta_t} \\
      &\le \sqrt{1-\zeta^2} G + (\sqrt{1-\zeta^2} + 1) \frac{\sqrt{1-\zeta^2}}{1 - \sqrt{1-\zeta^2}} G \\
      &\le \frac{(\sqrt{1-\zeta^2} - (1-\zeta^2)) + (1-\zeta^2 + \sqrt{1-\zeta^2})}{1 - \sqrt{1-\zeta^2}} G \\
      &\le \frac{2\sqrt{1-\zeta^2}}{1 - \sqrt{1-\zeta^2}} G \\
      &\le \frac{4}{\zeta^2} G.
    \end{align*}

    To conclude,
    \begin{align*}
      \min_{1 \le t \le T} F(x_t) - F(x^\star)
      &\le \frac{2R^2}{\sqrt T} \\
      &+ \frac 1T \sum_{t=0}^{T-1} \frac{1}{\sqrt{t+1}} \left( \frac{1}{2} \left(\frac{4}{\zeta^2}G\right)^2
      + \frac{2}{\zeta^2}G \frac{4}{\zeta^2}G + \frac{4}{\zeta^2} G^2 \right) \\
      &+ \frac{1}{2T^{\frac 32}} \frac{4}{\zeta^4}G^2 \\
      &\le \frac{2R^2}{\sqrt T} + \frac{2G^2}{\zeta^4 \sqrt T} \left( 20 + \frac{1}{T} \right),
    \end{align*}
    where we have used that \(\sum_{t=0}^{T-1} \frac{1}{\sqrt{t+1}} \le 2 \sqrt T\) and \(\frac{1}{\zeta^2} \le \frac{1}{\zeta^4}\).
  \end{proof}

	\section{Implementation details}
		\label{app:losses}
\neurocomp{
	This section provides detailed calculations for each step of Algorithm~\ref{alg:prox_meta} applied to Problem~\ref{opb:general} and for all losses presented in Table~\ref{tab:losses}.
}

\neurocomp{
	It is possible that a step has no closed-form expression but is the root of an equation.
	In this case (which is indicated by \(^\star\) below), the Newton-Raphson iteration is provided.
	In practice, less than \(10\) iterations of the Newton-Raphson method are enough to obtain a good approximation.
}

\neurocomp{
	For now on, let us note, for all $x\in\R$,
	$$
		\sign(x) = \begin{cases}
			-1 & \text{if } x<0 \\
			1 & \text{if } x>0 \\
			0 & \text{otherwise}.
		\end{cases}
	$$
}

\subsection{Least squares loss}
  \begin{description}
    \item[Definition:] $\ell(y, y') = (y-y')^2/2$.
    \item[Initial estimator:] $f_0 = \frac 1n \sum_{i=1}^n Y_i$.
    \item[Subgradient:] $\fsub C(f_t) = \left(\frac{f_t(X_i) - Y_i}{n} \right)_{1 \le i \le n}$.
    \item[Proximal direction:]
		\neurocomp{$\proxdir{\lambda} C(f_t) = \left( \frac{f_t(X_i) - Y_i}{\lambda+n} \right)_{1 \le i \le n}$.}
    \item[Line search:]
    $
      \gamma_{t+1} =
        \begin{cases}
          \frac{\sum_{i=1}^n (Y_i - f_t(X_i))g_{t+1}(X_i)}{\sum_{i=1}^n g_{t+1}(X_i)^2} & \text{if } \sum_{i=1}^n g_{t+1}(X_i)^2 > 0 \\
            0 & \text{otherwise}.
        \end{cases}
		$
  \end{description}

\subsection{Least absolute deviations loss}
  \begin{description}
    \item[Definition:] $\ell(y, y') = |y-y'|$.
    \item[Initial estimator:] $f_0$ is the empirical median of the sample $\{Y_1, \dots, Y_n\}$.
    \item[Subradient:] $\fsub C(f_t) = \left(\frac{\sign(f_t(X_i) - Y_i)}{n} \right)_{1 \le i \le n}$.
    \item[Proximal direction:]
		\neurocomp{
		\begin{align*}
			\proxdir{\lambda} C (f_t)
			&= \left( \frac{f_t(X_i) - Y_i}{\max \left( \lambda, n |f_t(X_i)-Y_i| \right)} \right)_{1 \le i \le n} \\
			&= \left( \begin{cases}
				\frac{\sign(f_t(X_i) - Y_i)}{n} & \text{if } |f_t(X_i)-Y_i| > \frac{\lambda}{n} \\
				\frac{f_t(X_i) - Y_i}{\lambda} & \text{otherwise}
			\end{cases} \right)_{1 \le i \le n}.
		\end{align*}
		}
    \item[Line search:]
    $ \gamma_{t+1} = \argmin_{\gamma \in \{0\} \cup \left\{ \frac{Y_i-f_t(X_i)}{g_{t+1}(X_i)} : g_{t+1}(X_i) \neq 0 \right\}} C(f_t + \gamma g_t)$.
  \end{description}

\subsection{Pinball loss}
  \begin{description}
    \item[Definition:] $\ell(y, y') = \max(\tau(y-y'), (\tau-1)(y-y'))$, $\tau \in (0, 1)$.
    \item[Initial estimator:] $f_0$ is the $\tau$-quantile of the sample $\{Y_1, \dots, Y_n\}$.
    \item[Subradient:]
    $
      \fsub C(f_t) = \left( \begin{cases}
        -\frac{\tau}{n} &\text{if } Y_i - f_t(X_i) > 0 \\
				\frac{1-\tau}{n} &\text{if } Y_i - f_t(X_i) < 0 \\
        0 &\text{otherwise} \\
      \end{cases} \right)_{1 \le i \le n}.
    $
    \item[Proximal direction:]
		\neurocomp{\[\proxdir{\lambda} C (f_t)
			= \left( \begin{cases}
				-\frac{\tau}{n} & \text{if } Y_i - f_t(X_i) > \frac{\lambda \tau}{n} \\
				\frac{1-\tau}{n} & \text{if } Y_i - f_t(X_i) < \frac{\lambda (\tau-1)}{n} \\
				\frac{f_t(X_i) - Y_i}{\lambda} & \text{otherwise}
		\end{cases} \right)_{1 \le i \le n}.\]}
    \item[Line search:]
    $ \gamma_{t+1} = \argmin_{\gamma \in \{0\} \cup \left\{ \frac{Y_i-f_t(X_i)}{g_{t+1}(X_i)} : g_{t+1}(X_i) \neq 0 \right\}} C(f_t + \gamma g_t)$.
  \end{description}

\subsection{Exponential loss}
  \begin{description}
    \item[Definition:] $\ell(y, y') = \exp(-\beta yy')$, $\beta>0$.
    \item[Initial estimator:] $f_0 = \frac{\log \left(\frac{p}{n-p} \right)}{2\beta}$, where $p = \sum_{1 \le i \le n \atop Y_i=1} 1$.
    \item[Subgradient:] $\fsub C(f_t) = \left(\frac{-\beta Y_i e^{-Y_i f_t(X_i)}}{n} \right)_{1 \le i \le n}$.
    \item[Proximal direction\(^\star\):]
		\neurocomp{$\proxdir{\lambda} C (f_t) = \left( \frac{f_t(X_i) - u_i}{\lambda} \right)_{1 \le i \le n}$,
		with Newton-\linebreak Raphson iteration \(u_i \gets u_i + \frac{f_t(X_i) - u_i + \frac{\lambda \beta Y_i}{n} \e^{-\beta Y_i u_i}}{1 + \frac{\lambda \beta^2}{n} \e^{-\beta Y_i u_i}}\).}
    \item[Line search\(^\star\):]
		\neurocomp{Newton-Raphson iteration \[\gamma_{t+1} \gets \gamma_{t+1} + \frac{\sum_{i=1}^n Y_i g_{t+1}(X_i) \e^{-\beta Y_i (f_t(X_i) + \gamma_{t+1} g_{t+1}(X_i))}}{\beta \sum_{i=1}^n g_{t+1}(X_i)^2 \e^{-\beta Y_i (f_t(X_i) + \gamma_{t+1} g_{t+1}(X_i))}}.\]}
  \end{description}

\subsection{Logistic loss}
  \begin{description}
    \item[Definition:] $\ell(y, y') = \log_2 (1+\exp(-yy'))$.
    \item[Initial estimator:] $f_0 = \log \left(\frac{p}{n-p} \right)$, where $p = \sum_{1 \le i \le n \atop Y_i=1} 1$.
    \item[Subgradient:] $\fsub C(f_t) = \left(\frac{-Y_i e^{-Y_i f_t(X_i)}}{n \log 2 (1 + e^{-Y_i f_t(X_i)})} \right)_{1 \le i \le n}$.
    \item[Proximal direction\(^\star\):]
		\neurocomp{$\proxdir{\lambda} C (f_t) = \left( \frac{f_t(X_i) - u_i}{\lambda} \right)_{1 \le i \le n}$,
		with Newton-\linebreak Raphson iteration \(u_i \gets u_i + \frac{f_t(X_i) - u_i + \frac{\lambda}{n \log 2}\frac{Y_i \e^{-Y_i u_i}}{1 + \e^{-Y_i u_i}}}{1 + \frac{\lambda}{n \log 2}\frac{\e^{-Y_i u_i}}{\left( 1 + \e^{-Y_i u_i} \right)^2}}\).}
    \item[Line search\(^\star\):]
    \neurocomp{Newton-Raphson iteration \[\gamma_{t+1} \gets \gamma_{t+1} + \frac{\sum_{i=1}^n \frac{Y_i g_{t+1}(X_i) \e^{-Y_i (f_t(X_i) + \gamma_{t+1} g_{t+1}(X_i))}}{1 + \e^{-Y_i (f_t(X_i) + \gamma_{t+1} g_{t+1}(X_i))}}}{\sum_{i=1}^n \frac{g_{t+1}(X_i)^2 \e^{-Y_i (f_t(X_i) + \gamma_{t+1} g_{t+1}(X_i))}}{\left( 1 + \e^{-Y_i (f_t(X_i) + \gamma_{t+1} g_{t+1}(X_i))} \right)^2}}.\]}
  \end{description}

\subsection{Hinge loss}
  \begin{description}
    \item[Definition:] $\ell(y, y') = \max(0, 1-yy')$.
    \item[Initial estimator:] $f_0 = \sign \left(\sum_{i=1}^n Y_i \right)$.
    \item[Subgradient:]
    $
      \fsub C(f_t) = \left( \begin{cases}
        -\frac{Y_i}{n} &\text{if } Y_if_t(X_i) < 1 \\
        0 &\text{otherwise} \\
      \end{cases} \right)_{1 \le i \le n}.
    $
    \item[Proximal direction:]
		\neurocomp{\[\proxdir{\lambda} C (f_t)
			= \left( \begin{cases}
				-\frac{Y_i}{n} & \text{if } Y_i f_t(X_i) < 1 - \frac{\lambda}{n} \\
				0 & \text{if } Y_i f_t(X_i) > 1 \\
				\frac{f_t(X_i) - Y_i}{\lambda} & \text{otherwise}
		\end{cases} \right)_{1 \le i \le n}.\]}
    \item[Line search:]
    $ \gamma_{t+1} = \argmin_{\gamma \in \{0\} \cup \left\{ \frac{1 - Y_if_t(X_i)}{Y_ig_{t+1}(X_i)} : g_{t+1}(X_i) \neq 0 \right\}} C(f_t + \gamma g_t)$.
  \end{description}

	\section{Accelerated proximal boosting in practice}
		\label{app:proof_weights}
Algorithm~\ref{alg:prox_acc_practice} describes a practical version of accelerated proximal boosting (Algorithm~\ref{alg:prox_acc}), which holds true also for accelerated gradient boosting \citep{biau_accelerated_2018}.
In accordance with the practice, the proximal steps are chosen adaptively by a line search (Line~\ref{lin:linesearch} of Algorithm~\ref{alg:prox_acc_practice}) and a shrinkage coefficient is introduced.


\begin{algorithm}[ht]
  \begin{algorithmic}[1]
    \REQUIRE
    $\nu \in (0, 1]$ (shrinkage coefficient), $\lambda > 0$ (proximal step).
    \STATE Set $g_0 \in \argmin_{g \in \F_0} C(g)$
    \COMMENT{initialization}.
    \STATE $x_0 \gets g_0(X_1^n) \in \R^n$
    \COMMENT{predictions}.
    \STATE $v_0 = x_0$
    \COMMENT{interpolated point}.
    \STATE $(w_0^{(0)}, \dots, w_T^{(0)}) \gets (1, 0, \dots, 0)$
    \COMMENT{weights of weak learners}.
    \FOR{$t=0$ \TO $T-1$}
      \STATE Compute \underline{(see Appendix~\ref{app:losses})}
      \[
        \begin{cases}
          r \gets -\fsub C (f_t) & \text{ for gradient boosting}, \\
          r \gets -\proxdir{\lambda} C(f_t) & \text{ for proximal boosting}.
        \end{cases}
      \]
      \STATE Compute \(g_{t+1} \in \argmin_{g \in \F} \norm{g(X_1^n) - r}\).
      \STATE Compute \(\gamma_{t+1} \in \argmin_{\gamma \in \R} C(f_t + \gamma g_{t+1})\)
      \underline{(see Appendix~\ref{app:losses})}. \label{lin:linesearch} 
      \STATE Set $x_{t+1} \gets v_t + \nu \gamma_{t+1} g_{t+1}(X_1^n)$
      \COMMENT{which corresponds to $x_{t+1} = f_{t+1}(X_1^n)$}. \label{lin:update2}
      \STATE Set $v_{t+1} \gets x_{t+1} + \alpha_{t+1}(x_{t+1} - x_t)$.
      \STATE Update weights $(w_0^{(t+1)}, \dots, w_{t+1}^{(t+1)})$ according to Property~\ref{prop:weights_rec}. \label{lin:update}
    \ENDFOR
    \ENSURE $f_T = \sum_{t=0}^T w_t^{(T)} g_t$.
  \end{algorithmic}
  \caption{Accelerated proximal/gradient boosting in practice.}
  \label{alg:prox_acc_practice}
\end{algorithm}

\label{sec:algo_practice}

As an additive model, it is of interest to express $f_T$ with respect to the base learners $(g_0, \dots, g_T)$ and their weights \(w_t\):
$f_T = \sum_{t=0}^T w_t g_t$.
For this purpose, the weights of the final model have to be tracked despite the recursive update of \(f_{t+1}\) (Line~\ref{lin:update3} in Algorithm~\ref{alg:prox_acc} and Line~\ref{lin:update2} in Algorithm~\ref{alg:prox_acc_practice}):
 \[
    f_{t+1} = f_t + \alpha_t(f_t - f_{t-1}) + \nu \gamma_{t+1} g_{t+1}.
\]
Property~\ref{prop:weights_final} gives the closed-form expression of the weights of $f_T$ in this case.

\begin{property}
  The weights of $f_T$ are:
  \[
    \begin{cases}
      w_0 = 1 \\
      w_1 = \nu \gamma_1 \\
      w_t = \left( 1 + \sum_{j=t}^{T-1} \prod_{k=t}^{j} \alpha_k \right) \nu \gamma_t, \forall t \in \{2,\dots, T-1\} \\
      w_T = \nu \gamma_T.
    \end{cases}
  \]
  \label{prop:weights_final}
\end{property}
\begin{proof}
  The update rule in Line~\ref{lin:update3} in Algorithm~\ref{alg:prox_acc} is:
  \[
    f_{t'+1} = (1 + \alpha_{t'}) f_{t'} - \alpha_{t'} f_{t'-1} + \nu \gamma_{t'+1} g_{t'+1},
  \]
  for all positive integers $t' \le T-1$.
  Let us denote, for each iteration $t' \in \{1, \dots, T-1\}$, $f_{t'} = \sum_{t=0}^{t'} w_t^{(t')} g_t$ the expansion of $f_{t'}$.
  Then
  \[
    f_{t'+1} = \sum_{t=0}^{t'-1} \left( (1 + \alpha_{t'})w_t^{(t')} - \alpha_{t'} w_t^{(t'-1)} \right) g_t  + (1 + \alpha_{t'}) w_{t'}^{(t')} g_{t'} + \nu \gamma_{t'+1} g_{t'+1}.
  \]
  First, we see that the weights of $g_{t'}$ and $g_{t'+1}$ in the expansion of $f_{t'+1}$ are respectively:
  \[
    \begin{cases}
      w_{t'}^{(t'+1)} = (1 + \alpha_{t'}) w_{t'}^{(t')} \\
      w_{t'+1}^{(t'+1)} = \nu \gamma_{t'+1}.
    \end{cases}
  \]
  Second, for each $t \in \{ 0, \dots, t'-1\}$, the weight of $g_t$ in the expansion of $f_{t'+1}$ is defined by:
  \[
    w_t^{(t'+1)} = (1 + \alpha_{t'})w_t^{(t')} - \alpha_{t'} w_t^{(t'-1)}.
  \]
  Therefore, considering that weights take value $0$ before being defined, \ie \linebreak $w_t^{(t-1)} = 0$, we have:
  \begin{align*}
    w_t^{(t'+1)} - w_t^{(t')}
    &= \alpha_{t'}(w_t^{(t')} - w_t^{(t'-1)}) \\
    &= \left( \prod_{k=t}^{t'} \alpha_k \right) (w_t^{(t)} - w_t^{(t-1)}) \\
    &= \left( \prod_{k=t}^{t'} \alpha_k \right) w_t^{(t)}.
  \end{align*}
  It follows that:
  \begin{align*}
    w_t^{(t'+1)}
    &= w_t^{(t')} + \left( \prod_{k=t}^{t'} \alpha_k \right) w_t^{(t)} \\
    &= w_t^{(t)} + \sum_{j=t}^{t'} \left( \prod_{k=t}^{j} \alpha_k \right) w_t^{(t)} \\
    &= \left( 1 + \sum_{j=t}^{t'} \prod_{k=t}^{j} \alpha_k \right)w_t^{(t)}.
  \end{align*}
  Then, for $k \le 1$, one has $\alpha_k = 0$, so $w_0^{(t'+1)} = w_0^{(0)} = 1$ and $w_1^{(t'+1)} = w_1^{(1)} = \nu \gamma_1$.
  Now, remarking that, for all $t \ge 2$, $w_t^{(t)} = \nu \gamma_t$, we can conclude that the weights of $f_T$ are:
  \[
    \begin{cases}
      w_0 = 1 \\
      w_1 = \nu \gamma_1 \\
      w_t = \left( 1 + \sum_{j=t}^{T-1} \prod_{k=t}^{j} \alpha_k \right) \nu \gamma_t, \quad \forall t \in \{2,\dots, T-1\} \\
      w_T = \nu \gamma_T.
    \end{cases}
  \]
\end{proof}

In addition, Property~\ref{prop:weights_rec} provides a recursive update suitable for implementing Algorithm~\ref{alg:prox_acc_practice}.
Let us remark that, Property~\ref{prop:weights_rec} is also valid for accelerated gradient boosting as proposed by \citet{biau_accelerated_2018}.
This paves the way to efficient implementations of both accelerated proximal and accelerated gradient boosting,
as done in the Python package \texttt{optboosing}\footnote{\url{https://github.com/msangnier/optboosting}}.

\begin{property}
  Let $f_{t} = \sum_{j=0}^{t} w_j^{(t)} g_j$ be the expansion of \(f_t\) at iteration \(t \in \{1, \dots, T-1\}\).
  Then, the weights can be updated according to the following recursion:
  \begin{equation}
    \begin{cases}
      w_0^{(0)} = 1 \\
      w_1^{(0)} = \nu \gamma_1 \\
      w_1^{(1)} = \nu \gamma_1 \\
      w_j^{(t+1)} = (w_j^{(t)} - w_j^{(t-1)}) (1 + \alpha_t) + w_j^{(t-1)}, \forall j \in \{1,\dots, t\} \\
      w_{t+1}^{(t+1)} = \nu \gamma_{t+1}.
    \end{cases}
    \label{equ:update}
  \end{equation}
  \label{prop:weights_rec}
\end{property}
\begin{proof}
  See proof of Property~\ref{prop:weights_final}.
\end{proof}

\end{document}